\documentclass{article}

\usepackage[nonatbib,final]{neurips_2019}

\usepackage[numbers]{natbib}

\usepackage[utf8]{inputenc} 
\usepackage[T1]{fontenc}    
\usepackage{hyperref}       
\usepackage{url}            
\usepackage{booktabs}       
\usepackage{amsfonts}       
\usepackage{nicefrac}       
\usepackage{microtype}      
\usepackage{amsmath}
\usepackage{amssymb}
\usepackage{mathtools}
\usepackage{float}
\usepackage{siunitx}

\usepackage{graphicx}
\usepackage[font=footnotesize]{caption}
\usepackage{subcaption}
\usepackage[noend]{algorithmic}
\usepackage{wrapfig}
\usepackage[ruled,vlined,noend, linesnumbered]{algorithm2e}
\usepackage{bm}
\usepackage{nccmath}
\usepackage{xcolor}
\usepackage{enumitem}
\usepackage{cleveref}
\usepackage[symbol]{footmisc}

\crefformat{section}{\S#2#1#3} 
\crefformat{subsection}{\S#2#1#3}
\crefformat{subsubsection}{\S#2#1#3}

\usepackage[titletoc, title]{appendix}

\DeclareMathOperator*{\argmax}{arg\,max}

\DeclarePairedDelimiterX{\infdivx}[2]{(}{)}{%
  #1\;\delimsize\|\;#2%
}
\newcommand{\infdiv}{D_\text{KL}\infdivx}

\newcommand{\state}{\mathbf{s}}
\newcommand{\action}{\mathbf{a}}
\newcommand{\task}{\mathcal{T}}
\newcommand{\experience}{\mathcal{D}_\task}
\newcommand{\traj}{\bm{\tau}}

\newcommand{\z}{\mathbf{z}}

\newcommand{\E}{\mathbb{E}}

\newcommand{\statespace}{\mathcal{S}}
\newcommand{\actionspace}{\mathcal{A}}
\newcommand{\reals}{\mathbb{R}}
\newcommand{\cmp}{\mathcal{C}}

\newcommand{\berk}{$\alpha$}
\newcommand{\toronto}{$\beta$}
\newcommand{\cmu}{$\gamma$}
\newcommand{\stanford}{$\delta$}

\newcommand{\method}{{CARML}}
\newcommand{\buffer}{\mathcal{D}}

\renewcommand{\thefootnote}{\fnsymbol{footnote}}
\title{Unsupervised Curricula \\ for Visual Meta-Reinforcement Learning}

\author{%
    Allan Jabri\textsuperscript{\normalfont \berk} \ Kyle Hsu\textsuperscript{\normalfont \toronto,$\dagger$} \ Benjamin Eysenbach\textsuperscript{\normalfont \cmu} \vspace{-100mm}
    \AND
    Abhishek Gupta\textsuperscript{\normalfont \berk} \ Sergey Levine\textsuperscript{\normalfont \berk} \ Chelsea Finn\textsuperscript{\normalfont \stanford} 
}

\begin{document}

\maketitle
{\let\thefootnote\relax\footnote{\hspace{-2.4mm}
\textsuperscript{\normalfont \berk}UC Berkeley \textsuperscript{\normalfont \toronto}University of Toronto \textsuperscript{\normalfont \cmu}Carnegie Mellon University \textsuperscript{\normalfont \stanford}Stanford University}}

\footnotetext[2]{Work done as a visiting student researcher at UC Berkeley.}

\vspace{-6mm}
\begin{abstract}
\vspace{-0.27cm}
In principle, meta-reinforcement learning algorithms leverage experience across many tasks to learn fast reinforcement learning (RL) strategies that transfer to similar tasks. However, current meta-RL approaches rely on manually-defined distributions of training tasks, and hand-crafting these task distributions can be challenging and time-consuming. Can ``useful'' pre-training tasks be discovered in an unsupervised manner? We develop an unsupervised algorithm for inducing an adaptive meta-training task distribution, i.e. an \textit{automatic curriculum}, by modeling unsupervised interaction in a visual environment. 
The task distribution is scaffolded by a parametric density model of the meta-learner's trajectory distribution. 
We formulate unsupervised meta-RL as information maximization between a latent task variable and the meta-learner’s data distribution, and describe a practical instantiation which alternates between integration of recent experience into the task distribution and meta-learning of the updated tasks. Repeating this procedure leads to iterative reorganization such that the curriculum adapts as the meta-learner's data distribution shifts. In particular, we show how discriminative clustering for visual representation can support trajectory-level task acquisition and exploration in domains with pixel observations, avoiding pitfalls of alternatives.
In experiments on vision-based navigation and manipulation domains, we show that the algorithm allows for unsupervised meta-learning that transfers to downstream tasks specified by hand-crafted reward functions and serves as pre-training for more efficient supervised meta-learning of test task distributions.
\end{abstract}

\vspace{-0.27cm}
\section{Introduction}
\vspace{-0.27cm}

The discrepancy between animals and learning machines in their capacity to gracefully adapt and {generalize} is a central issue in artificial intelligence research. The simple nematode \textit{C. elegans} is capable of adapting foraging strategies to varying scenarios \cite{calhoun2014maximally}, while many higher animals are driven to acquire reusable behaviors even without extrinsic task-specific rewards \citep{white1959motivation, piaget1954}. It is unlikely that we can build machines as adaptive as even the simplest of animals by exhaustively specifying shaped rewards or demonstrations across all possible environments and tasks. This has inspired work in reward-free learning~\citep{hastie2009unsupervised}, intrinsic motivation~\citep{singh2005intrinsically}, multi-task learning~\citep{rich1997multitask}, meta-learning~\citep{schmidhuber1987evolutionary}, and continual learning~\cite{thrun}.

An important aspect of generalization is the ability to share and transfer ability between related tasks.  In reinforcement learning (RL), a common strategy for multi-task learning is conditioning the policy on side-information related to the task. For instance, \textit{contextual} policies \cite{schaul2015universal} are conditioned on a task description (e.g. a \textit{goal}) that is meant to modulate the strategy enacted by the policy. Meta-learning of reinforcement learning (meta-RL) is yet more general as it places the burden of inferring the task on the learner itself, such that task descriptions can take a wider range of forms, the most general being an MDP.
In principle, meta-reinforcement learning (meta-RL) requires an agent to distill previous experience into fast and effective adaptation strategies for new, related tasks. However, the meta-RL framework by itself does not prescribe where this experience should come from; typically, meta-RL algorithms rely on being provided fixed, hand-specified task distributions, which can be tedious to specify for simple behaviors and intractable to design for complex ones \cite{hadfield2017inverse}. 
These issues beg the question of whether ``useful'' task distributions for meta-RL can be generated automatically.

\begin{figure}[t!]
    \centering
        \includegraphics[width=\textwidth]{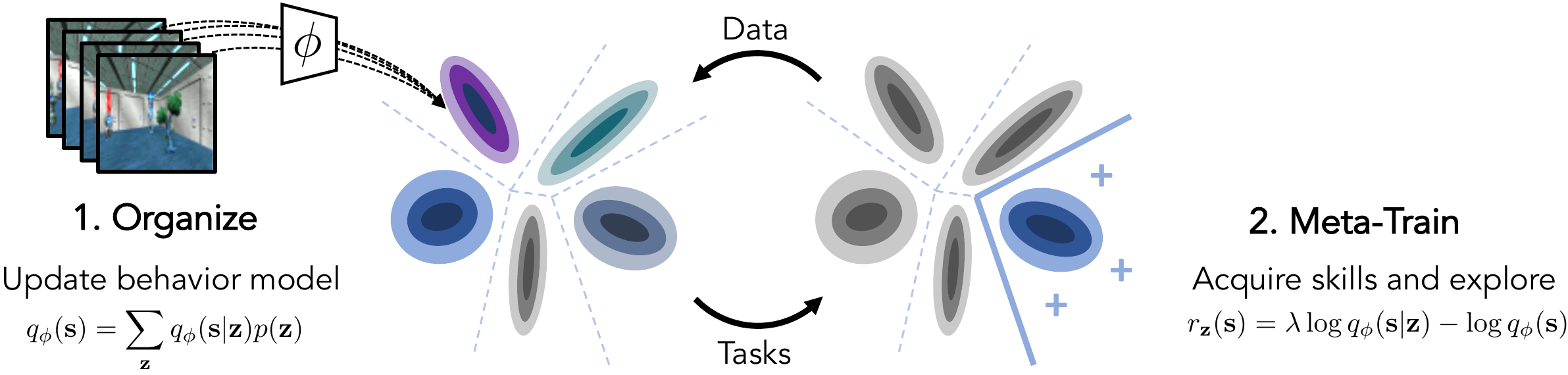}
        \vspace{-0.4cm}
    \caption{\footnotesize An illustration of \method{}, our approach for unsupervised meta-RL. We choose the behavior model $q_\phi$ to be a Gaussian mixture model in a jointly, discriminatively learned embedding space. An automatic curriculum arises from periodically re-organizing past experience via fitting $q_\phi$ and meta-learning an RL algorithm for performance over tasks specified using reward functions from $q_\phi$.}
    \label{fig:concept}
    \vspace{-5mm}
\end{figure}

In this work, we seek a procedure through which an agent in an environment with visual observations can automatically acquire useful (i.e. utility maximizing) behaviors, as well as how and when to apply them -- in effect allowing for \textit{unsupervised} pre-training in visual environments.
Two key aspects of this goal are: 1) learning to operationalize strategies so as to adapt to new tasks, i.e. meta-learning, and 2) unsupervised learning and exploration in the absence of explicitly specified tasks, i.e. skill acquisition \textit{without} supervised reward functions. These aspects interact insofar as the former implicitly relies on a task curriculum, while the latter is most effective when compelled by what the learner can and cannot do.
Prior work has offered a pipelined approach for unsupervised meta-RL consisting of unsupervised skill discovery followed by meta-learning of discovered skills, experimenting mainly in environments that expose low-dimensional ground truth state \cite{gupta2018unsupervised}. Yet, the aforementioned relation between skill acquisition and meta-learning suggests that they should not be treated separately. 

Here, we argue for closing the loop between skill acquisition and meta-learning in order to induce an \textit{adaptive} task distribution.
Such co-adaptation introduces a number of challenges related to the stability of learning and exploration. Most recent unsupervised skill acquisition approaches optimize for the discriminability of induced modes of behavior (i.e. \textit{skills}), typically expressing the discovery problem as a cooperative game between a policy and a learned reward function \cite{gregor2016variational, eysenbach2019diversity, achiam2018variational}. However, relying solely on discriminability becomes problematic in environments with high-dimensional (image-based) observation spaces as it results in an issue akin to mode-collapse in the task space. This problem is further complicated in the setting we propose to study, wherein the policy data distribution is that of a meta-learner rather than a contextual policy. We will see that this can be ameliorated by specifying a hybrid discriminative-generative model for parameterizing the task distribution.

The main contribution of this paper is an approach for inducing a task curriculum for unsupervised meta-RL in a manner that scales to domains with pixel observations. Through the lens of information maximization, we frame our unsupervised meta-RL approach as variational expectation-maximization (EM), in which the E-step corresponds to fitting a task distribution to a meta-learner's behavior and the M-step to meta-RL on the current task distribution with reinforcement for both skill acquisition and exploration.
For the E-step, we show how deep discriminative clustering allows for trajectory-level representations suitable for learning diverse skills from pixel observations.
Through experiments in vision-based navigation and robotic control domains, we demonstrate that the approach i) enables an unsupervised meta-learner to discover and meta-learn skills that transfer to downstream tasks specified by human-provided reward functions, and ii) can serve as pre-training for more efficient supervised meta-reinforcement learning of downstream task distributions. 

\vspace{-0.27cm}
\section{Preliminaries: Meta-Reinforcement Learning} \label{sec:prelims}
\vspace{-0.27cm}
\textit{Supervised} meta-RL optimizes an RL algorithm $f_\theta$ for performance on a hand-crafted distribution of tasks $p(\task)$, where $f_\theta$ might take the form of an recurrent neural network (RNN) implementing a learning algorithm~\citep{duan2016rl,wang2016learning}, or a function implementing a gradient-based learning algorithm~\cite{finn2017model}. 
Tasks are Markov decision processes (MDPs) \mbox{$\task_i = (\statespace, \actionspace, r_i, P, \gamma, \rho, T)$} consisting of state space $\statespace$, action space $\actionspace$, reward function \mbox{$r_i: \statespace \times \actionspace \to \reals$}, probabilistic transition dynamics $P(\state_{t+1}|\state_t,\action_t)$, discount factor $\gamma$, initial state distribution $\rho(\state_1)$, and finite horizon $T$.  Often, and in our setting, tasks are assumed to share $\statespace, \actionspace$.
For a given $\task \sim p(\task)$, $f_\theta$ learns a policy $\pi_\theta(\action|\state,\experience)$ conditioned on task-specific experience. Thus, a meta-RL algorithm optimizes $f_\theta$ for expected performance of $\pi_\theta(\action|\state,\experience)$ over $p(\task)$, such that it can generalize to unseen test tasks also sampled from $p(\task)$.

For example, RL$^2$~\citep{duan2016rl,wang2016learning} chooses $f_\theta$ to be an RNN with weights $\theta$. For a given task $\task$, $f_\theta$ hones $\pi_\theta(\action | \state, \experience)$ as it recurrently ingests \mbox{$\experience = (\state_1, \action_1, r(\state_1, \action_1), d_1, \dots)$}, the sequence of states, actions, and rewards produced via interaction within the MDP.
Crucially, the same task is seen several times, and the hidden state is not reset until the next task.
The loss is the negative discounted return obtained by $\pi_\theta$ across episodes of the same task, and $f_\theta$ can be optimized via standard policy gradient methods for RL, backpropagating gradients through time and across episode boundaries.

\emph{Unsupervised} meta-RL aims to break the reliance of the meta-learner on an explicit, upfront specification of $p(\task)$. Following \citet{gupta2018unsupervised}, we consider a controlled Markov process (CMP) $\cmp=(\statespace, \actionspace, P, \gamma, \rho, T)$, which is an MDP without a reward function. We are interested in the problem of learning an RL algorithm $f_\theta$ via unsupervised interaction within the CMP such that once a reward function $r$ is specified at test-time, $f_\theta$ can be readily applied to the resulting MDP to efficiently maximize the expected discounted return.

Prior work~\cite{gupta2018unsupervised} pipelines skill acquisition and meta-learning by pairing an unsupervised RL algorithm DIAYN \citep{eysenbach2019diversity} and a meta-learning algorithm MAML \citep{finn2017model}: first, a contextual policy is used to discover skills in the CMP, yielding a finite set of learned reward functions distributed as $p(r)$;
then, the CMP is combined with a frozen $p(r)$ to yield $p(\task)$, which is fed to MAML to meta-learn $f_\theta$. 
In the next section, we describe how we can generalize and improve upon this pipelined approach by jointly performing skill acquisition as the meta-learner learns and explores in the environment. 

\begin{figure}
  \begin{minipage}[c]{0.6\textwidth}
    \vspace{-5mm}
    \includegraphics[width=\textwidth]{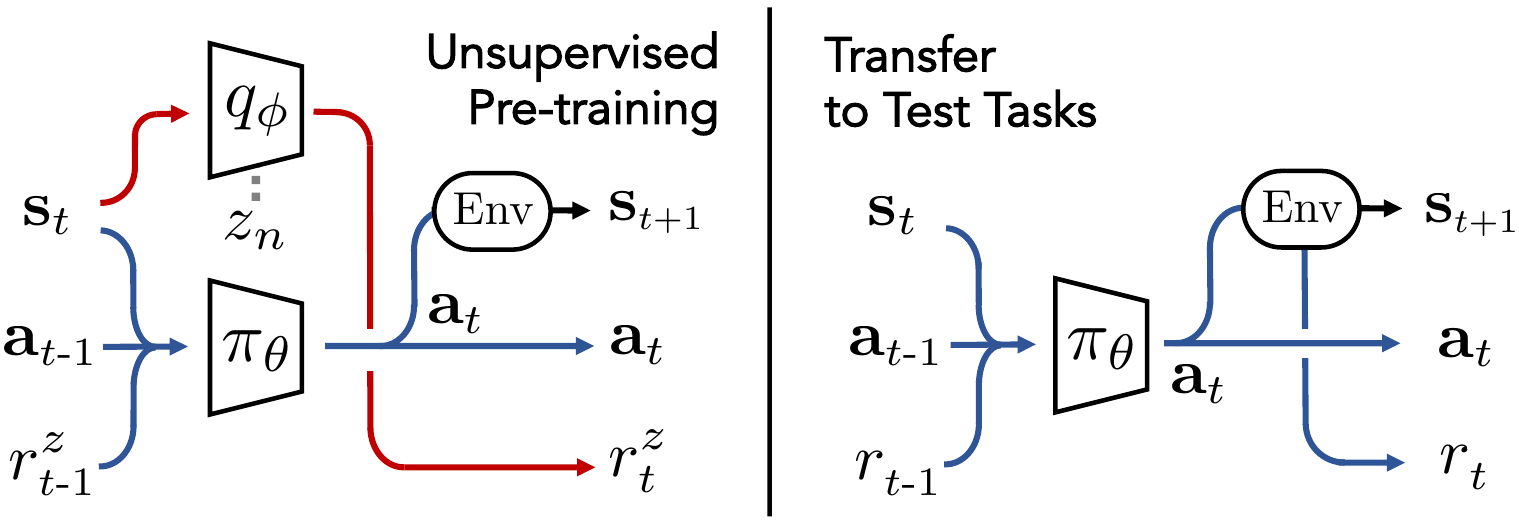}
  \end{minipage}\hfill
  \begin{minipage}[c]{0.38\textwidth}
    \caption{ A step for the meta-learner.  \textbf{(Left) Unsupervised pre-training.} The policy meta-learns self-generated tasks based on the behavior model $q_\phi$.  \textbf{ (Right) Transfer.} Faced with new tasks, the policy transfers acquired meta-learning strategies to maximize unseen reward functions.}
    \label{fig:train-test}
  \end{minipage}
  \vspace{-8mm}
\end{figure}

\vspace{-0.27cm}
\section{\underline{C}urricul\underline{a} for Unsupervised Meta-\underline{R}einforce\underline{m}ent \underline{L}earning} \label{sec:method}
\vspace{-0.27cm}

Meta-learning is intended to prepare an agent to efficiently solve new tasks related to those seen previously. To this end, the meta-RL agent must balance 1) exploring the environment to infer which task it should solve, and 2) visiting states that maximize reward under the inferred task. The duty of unsupervised meta-RL is thus to present the meta-learner with tasks that allow it to practice task inference and execution, without the need for human-specified task distributions. Ideally, the task distribution should exhibit both structure and diversity. That is, the tasks should be distinguishable and not excessively challenging so that a developing meta-learner can infer and execute the right skill, but, for the sake of generalization, they should also encompass a diverse range of associated stimuli and rewards, including some beyond the current scope of the meta-learner. 
Our aim is to strike this balance by inducing an adaptive task distribution.

With this motivation, we develop an algorithm for unsupervised meta-reinforcement learning in visual environments that constructs a task distribution without supervision. The task distribution is derived from a latent-variable density model of the meta-learner's cumulative behavior, with exploration based on the density model driving the evolution of the task distribution. As depicted in Figure\ref{fig:concept}, learning proceeds by alternating between two steps: \textbf{organizing experiential data} (i.e., trajectories generated by the meta-learner) by modeling it with a mixture of latent components forming the basis of ``skills'', and {meta-reinforcement learning} by \textbf{treating these skills as a training task distribution}. 

Learning the task distribution in a data-driven manner ensures that tasks are feasible in the environment. 
While the induced task distribution is in no way guaranteed to align with test task distributions, it may yet require an implicit understanding of structure in the environment. This can indeed be seen from our visualizations in~\cref{sec:experiments}, which demonstrate that acquired tasks show useful structure, though in some settings this structure is easier to meta-learn than others.
In the following, we formalize our approach, \method{}, through the lens of information maximization and describe a concrete instantiation that scales to the vision-based environments considered in \cref{sec:experiments}. 

\vspace{-0.2cm}
\subsection{An Overview of \method{}} \label{sec:overview}
\vspace{-0.2cm}

We begin from the principle of information maximization (IM), which has been applied across unsupervised representation learning \cite{bell_sejnowski_1995, barber2004im, oord2018representation} and reinforcement learning \cite{mohamed15, gregor2016variational} for organization of data involving latent variables. In what follows, we organize data from our policy by maximizing the mutual information (MI) between state trajectories $\traj:= (\state_1,\dots,\state_T)$ and a latent task variable $\z$.
This objective provides a principled manner of trading-off structure and diversity:
from ${I}(\traj; \z) := H(\traj) - H(\traj | \z)$, we see that $H(\traj)$ promotes coverage in policy data space (i.e. \textit{diversity}) while $-H(\traj|\z)$ encourages a lack of diversity under each task (i.e. \textit{structure} that eases task inference).

We approach maximizing ${I}(\traj; \z)$ exhibited by the meta-learner $f_\theta$ via variational EM~\citep{barber2004im}, introducing a
variational distribution $q_\phi$ that can intuitively be viewed as a task scaffold for the meta-learner. 
In the E-step, we fit $q_\phi$ to a reservoir of trajectories produced by $f_\theta$, re-organizing the cumulative experience. In turn, $q_\phi$ gives rise to a task distribution $p(\task)$: each realization of the latent variable $\z$ induces a reward function $r_\z (\state)$, which we combine with the CMP $\cmp_i$ to produce an MDP $\task_i$ (Line~\ref{line:reward}). In the M-step, $f_\theta$ meta-learns the task distribution $p(\task)$. Repeating these steps forms a curriculum in which the task distribution and meta-learner co-adapt: each M-step adapts the meta-learner $f_\theta$ to the updated task distribution, while each E-step updates the task scaffold  $q_\phi$ based on the data collected during meta-training. Pseudocode for our method is presented in Algorithm~\ref{alg:ours}.

\vspace{-2mm}
\begin{algorithm}[h!]
  \caption{\method{}}
  \label{alg:ours}
\begin{algorithmic}[1]
\footnotesize
  \STATE {\bfseries Require:} $\cmp$, an MDP without a reward function
  \STATE Initialize $f_\theta$, an RL algorithm parameterized by $\theta$.
  \STATE Initialize $\buffer$, a reservoir of state trajectories, via a randomly initialized policy.
 
  \WHILE{not done}
    \STATE Fit a task-scaffold $q_\phi$ to $\buffer$, e.g. by using Algorithm~\ref{alg:em}. \hfill \textbf{E-step \cref{sec:e-step}} \label{line:fit}
    \FOR{a desired mixture model-fitting period}
      \STATE Sample a latent task variable $\z \sim q_\phi(\z)$.
      \STATE Define the reward function $r_\z(\state)$, e.g. by Eq.~\ref{eq:gen_reward}, and a task $\task = \cmp \cup r_\z(\state)$. \label{line:reward}
      \STATE Apply $f_\theta$ on task $\task$ to obtain a policy $\pi_\theta(\action | \state, \experience)$ and trajectories $\{\traj_i\}$.
      \STATE Update $f_\theta$ via a meta-RL algorithm, e.g. RL$^2$~\citep{duan2016rl}. \hfill \textbf{M-step \cref{sec:m-step}}
      \STATE Add the new trajectories to the reservoir: $\buffer \gets \buffer \cup \{\traj_i\}$.
    \ENDFOR
  \ENDWHILE
  
  \STATE {\bfseries Return:} a meta-learned RL algorithm $f_\theta$ tailored to $\cmp$
\end{algorithmic}
\end{algorithm}

\vspace{-5mm}
\subsection{E-Step: Task Acquisition} \label{sec:e-step}
\vspace{-0.2cm}
The purpose of the E-step is to update the task distribution by integrating changes in the meta-learner's data distribution with previous experience, thereby allowing for re-organization of the task scaffold. This data is from the \textit{post-update} policy, meaning that it comes from a policy $\pi_\theta(\action|\state,\experience)$ conditioned on data collected by the meta-learner for the respective task. In the following, we abuse notation by writing $\pi_\theta(\action|\state,\z)$ -- conditioning on the latent task variable $\z$ rather than the task experience $\experience$.

The general strategy followed by recent approaches for skill discovery based on IM is to lower bound the objective by introducing a variational posterior $q_\phi(\z|\state)$ in the form of a classifier. In these approaches, the E-step  amounts to updating the classifier to discriminate between data produced by different skills as much as possible. A potential failure mode of such an approach is an issue akin to mode-collapse in the task distribution, wherein the policy drops modes of behavior to favor easily discriminable trajectories, resulting in a lack of diversity in the task distribution and no incentive for exploration; this is especially problematic when considering high-dimensional observations. Instead, here we derive a generative variant, which allows us to account for explicitly capturing modes of behavior (by optimizing for likelihood), as well as a direct mechanism for exploration. 

We introduce a variational distribution $q_\phi$, which could be e.g. a (deep) mixture model with discrete $\z$ or a variational autoencoder (VAE)~\citep{kingma2014auto} with continuous $\z$, lower-bounding the objective:

\vspace{-1mm}
\begin{align}
    I(\traj;\z) 
    &= -\sum_{\traj} \pi_\theta(\traj) \log \pi_\theta(\traj) + \sum_{\traj,\z} \pi_\theta(\traj, \z) \log \pi_\theta(\traj|\z) \\
    &\geq -\sum_{\traj} \pi_\theta(\traj) \log \pi_{\theta}(\traj) + \sum_{\traj,\z} \pi_\theta(\traj|\z) q_\phi(\z) \log q_\phi(\traj|\z) \label{eq:gen_variational} 
\end{align}
\vspace{-4mm}

The E-step corresponds to optimizing Eq.~\ref{eq:gen_variational} with respect to $\phi$, and thus amounts to fitting $q_\phi$ to a reservoir of trajectories $\buffer{}$ produced by $\pi_\theta$:

\begin{equation}
   \max_{\phi} \; \E_{\z \sim q_\phi(\z), \traj \sim \buffer } 
   \big[ 
        \log q_\phi(\traj|\z) 
   \big]
\end{equation}
\vspace{-3mm}

What remains is to determine the form of $q_\phi$.
We choose the variational distribution to be a state-level mixture density model \mbox{$q_\phi(\state,\z)= q_\phi(\state|\z)q_\phi(\z)$}. 
Despite using a state-level generative model, we can treat $\z$ as a trajectory-level latent by computing the trajectory-level likelihood as the factorized product of state likelihoods (Algorithm~\ref{alg:em}, Line 4). This is useful for obtaining trajectory-level tasks; in the  M-step (\cref{sec:m-step}), we map samples from $q_\phi(\z)$ to reward functions to define tasks for meta-learning. 

\noindent\begin{minipage}[t]{.6\textwidth}
\begin{algorithm}[H]
	\caption{Task Acquisition via Discriminative Clustering}
	\label{alg:em}
	\begin{algorithmic}[1]
		\STATE {\bfseries Require:} a set of trajectories $\buffer=\{(\state_1,\dots,\state_T)\}_{i=1}^N$
		\STATE Initialize $(\phi_w, \phi_m)$, encoder and mixture parameters.
		\WHILE{not converged}
		\STATE{Compute $L(\phi_m;\traj, z) = \sum_{\state_t \in \traj} \log q_{\phi_m}(g_{\phi_w}(\state_t) | z)$. \label{line:aggregate}} 
		\vspace{-3mm}
		\STATE{$\phi_m \gets \argmax_{\phi'_m}\sum_{i=1}^N L(\phi'_m;\traj_i,z)$ (via MLE)} 
		\STATE{$\mathcal{D} := \{(\state, y:=\argmax_k q_{\phi_m}(z=k | g_{\phi_w}(\state))\}$.}
		\STATE{$\phi_w \gets \argmax_{\phi'_w} \sum_{(\state, y) \in \mathcal{D}} \log q(y | g_{\phi'_w}(\state))$}
		\ENDWHILE
		\STATE {\bfseries Return:} a mixture model $q_\phi(\state,z)$
	\end{algorithmic}
\end{algorithm}

\end{minipage} \hspace{2mm}
\begin{minipage}[t]{.35\textwidth}
  \centering\
    \vspace{-5mm}
    \begin{figure}[H]
    \centering
    \includegraphics[width=0.98\textwidth]{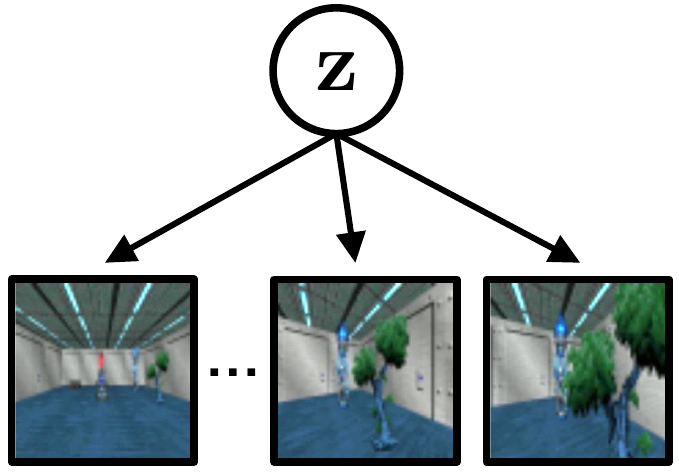}
    \vspace{-3pt}
    \caption{\footnotesize
    Conditional independence \\ assumption for states along a trajectory. 
        }
    \label{fig:graphical_model}
\end{figure}
\end{minipage}

\textbf{Modeling Trajectories of Pixel Observations.}
While models like the variational autoencoder have been used in related settings~\citep{nair2018visual}, a basic issue is that optimizing for reconstruction treats all pixels equally. We, rather, will tolerate \textit{lossy} representations as long as they capture \textit{discriminative} features useful for stimulus-reward association.
Drawing inspiration from recent work on unsupervised feature learning by clustering \cite{bojanowski2017unsupervised, caron2018deep}, we propose to fit the trajectory-level mixture model via discriminative clustering, striking a balance between discriminative and generative approaches.

We adopt the optimization scheme of DeepCluster~\cite{caron2018deep}, which alternates between i) clustering representations to obtain pseudo-labels and ii) updating the representation by supervised learning of pseudo-labels. In particular, we derive a trajectory-level variant (Algorithm~\ref{alg:em}) by forcing the responsibilities of all observations in a trajectory to be the same (see Appendix~\ref{app:traj_derivation} for a derivation), leading to state-level visual representations optimized with trajectory-level supervision.

The conditional independence assumption in Algorithm~\ref{alg:em} is a simplification insofar as it discards the order of states in a trajectory. 
However, if the dynamics exhibit continuity and causality, the visual representation might yet capture temporal structure, since, for example, attaining certain observations might imply certain antecedent subtrajectories. We hypothesize that a state-level model can regulate issues of over-expressive sequence encoders, which have been found to lead to skills with undesirable attention to details in dynamics~\citep{achiam2018variational}.
As we will see in~\cref{sec:experiments}, learning representations under this assumption still
allows for learning visual features that capture trajectory-level structure. 

\vspace{-0.2cm}
\subsection{M-Step: Meta-Learning}\label{sec:m-step}
\begin{figure}[H]
  \begin{minipage}[c]{0.6\textwidth}
    \vspace{-5mm}
    \includegraphics[width=\textwidth,]{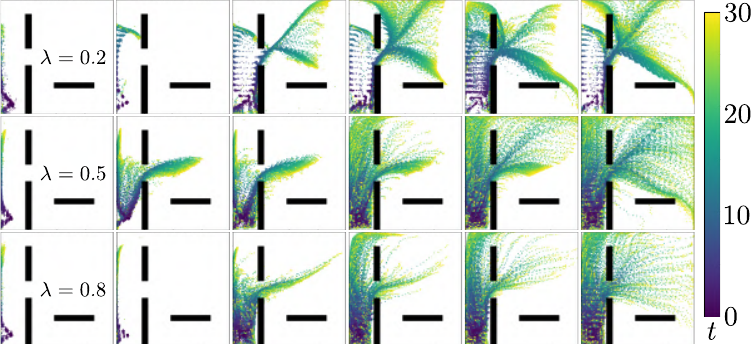}
  \end{minipage}\hspace{2mm}
  \begin{minipage}[c]{0.38\textwidth}
    \caption{\footnotesize Balancing consistency and exploration with $\lambda$ in a simple 2D maze environment. Each row shows a progression of tasks developed over the course of training. Each box presents the mean reconstructions under a VAE $q_\phi$ (Appendix~\ref{app:sec:vae}) of 2048 trajectories. Varying $\lambda$ of Eq.~\ref{eq:gen_reward} across rows, we observe that a small $\lambda$ (top) results in aggressive exploration; a large $\lambda$ (bottom) yields relatively conservative behavior; and a moderate $\lambda$ (middle) produces sufficient exploration and a smooth task distribution.}
    \label{fig:lambda}
  \end{minipage}
  \vspace{-4mm}
\end{figure}

\vspace{-0.2cm}
Using the task scaffold updated via the E-step, we meta-learn $f_\theta$ in the M-step so that $\pi_\theta$ can be quickly adapted to tasks drawn from the task scaffold. To define the task distribution, we must specify a form for the reward functions $r_\z(\state)$.
To allow for state-conditioned Markovian rewards rather than non-Markovian trajectory-level rewards, we lower-bound the trajectory-level MI objective:
\vspace{-0.1cm}
\begin{align}
    {I}(\traj; \z) &= 
    \frac{1}{T} \sum_{t=1}^T {H}(\z) - {H}(\z|\state_{1}, ..., \state_{T}) 
    \geq \frac{1}{T} \sum_{t=1}^T {H}(\z) - {H}(\z|\state_t) 
    \\
     &\geq \E_{\z \sim q_\phi(\z), \state \sim \pi_\theta(\state | \z)} 
     \big[
        \log q_\phi(\state | \z) - \log \pi_\theta(\state)
    \big] \label{eq:state_variational_MI}
\end{align}
We would like to optimize the meta-learner under the variational objective in Eq.~\ref{eq:state_variational_MI},
but optimizing the second term, the policy's state entropy, is in general intractable. Thus, we make the simplifying assumption that the fitted variational marginal distribution matches that of the policy:
 \begin{align} 
    &\max_{\theta} \; \E_{\z \sim q_\phi(\z), \state \sim \pi_\theta(\state | \z)} \big[\log q_\phi(\state | \z) - \log q_\phi(\state) \big] \label{eq:gen_pol} \\
    =&\max_{\theta} \; I(\pi_\theta(\state); q_\phi(\z)) -\infdiv{\pi_\theta(\state|\z)}{q_\phi(\state|\z)} + \infdiv{\pi_\theta(\state)}{q_\phi(\state))}  \label{eq:gen_info}
\end{align}
Optimizing Eq.~\ref{eq:gen_pol} amounts to maximizing the reward of \mbox{$r_\z(\state) = \log q_\phi(\state|\z) - \log q_\phi(\state)$}. As shown in Eq.~\ref{eq:gen_info}, this corresponds to information maximization between the policy's state marginal and the latent task variable, along with terms for matching the task-specific policy data distribution to the corresponding mixture mode and deviating from the mixture's marginal density. 
We can trade-off between component-matching and exploration by introducing a weighting term $\lambda \in [0, 1]$ into $r_\z(\state)$:

\vspace{-2mm}
\begin{align}
    r_\z(\state) &= \lambda \log q_\phi(\state|\z) - \log q_\phi(\state) \label{eq:gen_reward} \\
    &= (\lambda - 1)  \log  q_\phi(\state|\z) + \log q_\phi(\z|\state) + C \label{eq:gen_reward3}
\end{align}
where $C$ is a constant with respect to the optimization of $\theta$. From Eq.~\ref{eq:gen_reward3}, we can interpret $\lambda$ as trading off between discriminability of skills and task-specific exploration. Figure~\ref{fig:lambda} shows the effect of tuning $\lambda$ on the structure-diversity trade-off alluded to at the beginning of~\cref{sec:method}.

\vspace{-8mm}
\section{Related Work}
\label{sec:related-work}
\vspace{-0.27cm}

\textbf{Unsupervised Reinforcement Learning}. 
Unsupervised learning in the context of RL is the problem of enabling an agent to learn about its environment and acquire useful behaviors without human-specified reward functions. 
A large body of prior work has studied exploration and intrinsic motivation objectives~\citep{schmidhuber2009driven,salge2014empowerment,pathak2017curiosity,fu2017ex2,burda2019exploration,bellemare2016unifying,lehman2011abandoning,osband2018randomized}.
These algorithms do not aim to acquire skills that can be operationalized to solve tasks, but rather try to achieve wide coverage of the state space; our objective (Eq.~\ref{eq:gen_reward}) reduces to pure density-based exploration with $\lambda = 0$. Hence, these algorithms still rely on slow RL~\citep{botvinick2019reinforcement} in order to adapt to new tasks posed at test-time. 
Some prior works consider unsupervised pre-training for efficient RL, but these works typically focus on settings in which exploration is not as much of a challenge~\citep{watter2015embed, finn2017deep, ebert2017self}, focus on goal-conditioned policies \cite{pathak2018zero, nair2018visual}, or have not been shown to scale to high-dimensional visual observation spaces~\citep{lopes2012exploration,shyam2019model}. 
Perhaps most relevant to our work are unsupervised RL algorithms for learning reward functions via optimizing information-theoretic objectives involving latent skill variables~\citep{gregor2016variational,achiam2018variational,eysenbach2019diversity,warde2019unsupervised}.
In particular, with a choice of $\lambda = 1$ in Eq.~\ref{eq:gen_reward3} we recover the information maximization objective used in prior work~\citep{achiam2018variational, eysenbach2019diversity}, besides the fact that we simulatenously perform meta-learning.
The setting of training a contextual policy with a classifier as $q_\phi$ in our proposed framework (see Appendix~\ref{app:sec:diayn}) provides an interpretation of DIAYN as implicitly doing trajectory-level clustering. \citet{warde2019unsupervised} also considers accumulation of tasks, but with a focus on goal-reaching  and by maintaining a goal reservoir via heuristics that promote diversity.

\textbf{Meta-Learning}. Our work is distinct from above works in that it formulates a meta-learning approach to explicitly train, without supervision, for the ability to adapt to new downstream RL tasks. Prior work \citep{hsu2019unsupervised, khodadadeh2018unsupervised, antoniou2019assume} 
has investigated this unsupervised meta-learning setting for image classification; the setting considered herein is complicated by the added challenges of RL-based policy optimization and exploration. \citet{gupta2018unsupervised} provides an initial exploration of the unsupervised meta-RL problem, proposing a straightforward combination of unsupervised skill acquisition (via DIAYN) followed by MAML \cite{finn2017model} with experiments restricted to environments with fully observed, lower-dimensional state. Unlike these works and other meta-RL works~\citep{wang2016learning,duan2016rl,mishra2018simple,rakelly2019efficient,finn2017model,houthooft2018evolved,gupta2018meta,rothfuss2019promp,stadie2018some,sung2017learning}, we close the loop to jointly perform task acquisition and meta-learning so as to achieve an automatic curriculum to facilitate joint meta-learning and task-level exploration.

\textbf{Automatic Curricula}. The idea of automatic curricula has been widely explored both in supervised learning and RL. In supervised learning, interest in automatic curricula is based on the hypothesis that exposure to data in a specific order (i.e. a non-uniform curriculum) may allow for learning harder tasks more efficiently \cite{elman1993learning, schmidhuber2009driven,graves2017automated}. 
In RL, an additional challenge is exploration; hence, related work in RL considers the problem of \emph{curriculum generation}, whereby the task distribution is designed to guide exploration towards solving complex tasks \cite{florensa2017reverse, matiisen2019teacher, florensa2017automatic, schmidhuber2011powerplay} or unsupervised pre-training \cite{sukhbaatar2018intrinsic, forestier2017intrinsically}. Our work is driven by similar motivations, though we consider a curriculum in the setting of meta-RL and frame our approach as information maximization.

\section{Experiments}\label{sec:experiments}
\vspace{-0.27cm}

We experiment in visual navigation and visuomotor control domains to study the following questions:
\vspace{-6mm}
\begin{itemize}[leftmargin=10mm]
  \setlength\itemsep{-0.1em}
    \item What kind of tasks are discovered through our task acquisition process (the E-step)? 
    \item Do these tasks allow for meta-training of strategies that transfer to test tasks?
    \item Does closing the loop to jointly perform task acquisition and meta-learning bring benefits? 
    \item Does pre-training with \method{} accelerate meta-learning of test task distributions?
\end{itemize}
\vspace{-8pt}
Videos are available at the project website \url{https://sites.google.com/view/carml}.

\vspace{-3mm}
\subsection{Experimental Setting}
\vspace{-0.2cm}

The following experimental details are common to the two vision-based environments we consider. Other experimental  are explained in more detail in Appendix~\ref{app:sec:implementation_details}. \vspace{-1mm}

\textbf{Meta-RL.} \method{} is agnostic to the meta-RL algorithm used in the M-step. We use the RL$^2$ algorithm \citep{duan2016rl}, which has previously been evaluated on simpler visual meta-RL domains, with a PPO~\citep{schulman2017ppo} optimizer. Unless otherwise stated, we use four episodes per trial (compared to the two episodes per trial used in \cite{duan2016rl}), since the settings we consider involve more challenging task inference.

\vspace{-1mm}
\textbf{Baselines.} We compare against: 1) PPO from scratch on each evaluation task, 2) pre-training with random network distillation (RND)~\citep{burda2019exploration} for unsupervised exploration, followed by fine-tuning on evaluation tasks, and 3) supervised meta-learning on the test-time task distribution, as an oracle. 

\vspace{-1mm}
\textbf{Variants.} We consider variants of our method to ablate the role of design decisions related to task acquisition and joint training: 
    4) \textit{pipelined} (most similar to \cite{gupta2018unsupervised}) -- task acquisition with a contextual policy, followed by meta-RL with RL$^2$;
    5) \textit{online discriminator} -- task acquisition with a purely discriminative $q_\phi$ (akin to online DIAYN);
    and 6) \textit{online pretrained-discriminator} -- task acquisition with a discriminative $q_\phi$ initialized with visual features trained via Algorithm~\ref{alg:em}.
\vspace{-0.2cm}
\subsection{Visual Navigation}
\label{sec:viznav}
\vspace{-0.2cm}

The first domain we consider is first-person visual navigation in ViZDoom \cite{kempka2016vizdoom}, involving a room filled with five different objects (drawn from a set of 50). We consider a setup akin to those featured in \cite{chaplot2018gated, xie2018few} (see Figure~\ref{fig:graphical_model}). 
The true state consists of continuous 2D position and continuous orientation, while observations are egocentric images with limited field of view. Three discrete actions allow for turning right or left, and moving forward.
We consider two ways of sampling the CMP $\mathcal{C}$.
\textbf{Fixed}: fix a set of five objects and positions for both unsupervised meta-training and testing. \textbf{Random}: sample five objects and randomly place them (thereby randomizing the state space and dynamics).

\begin{figure}[b!]
    \vspace{-5mm}
    \centering\includegraphics[width=0.99\textwidth]{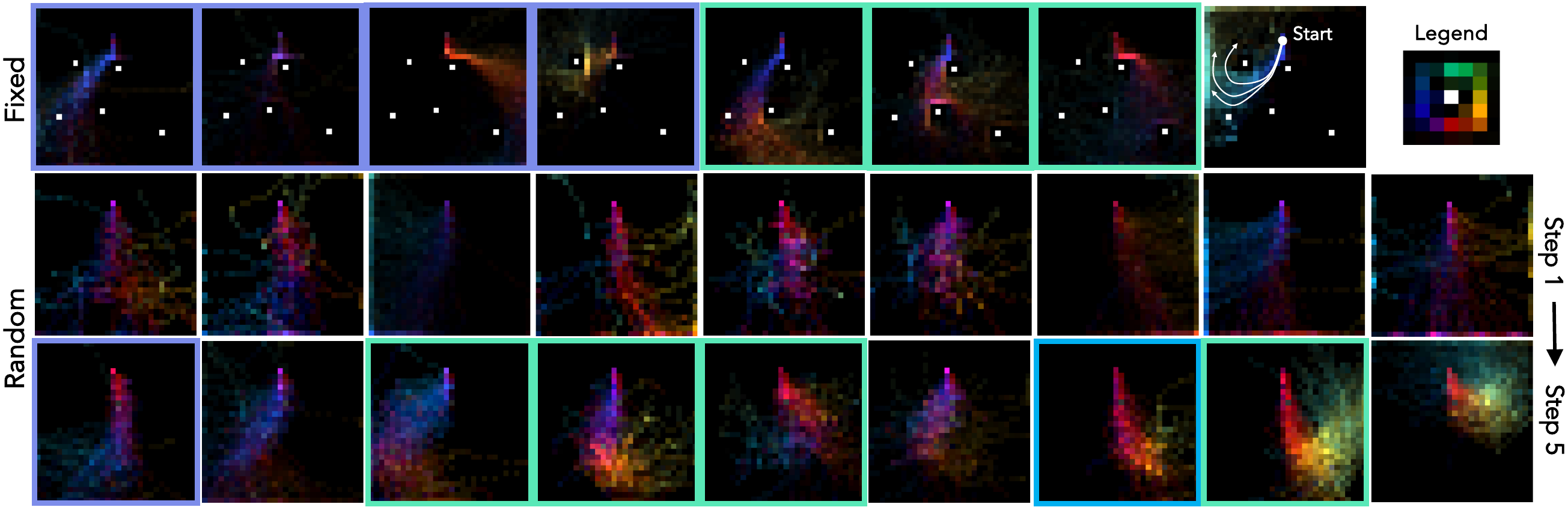}
    \vspace{-1mm}
    \caption{Skill maps for visual navigation. We visualize some of the discovered tasks by projecting trajectories of certain mixture components into the true state space. White dots correspond to fixed objects. The legend indicates orientation as color; on its left is an interpretation of the depicted component. Some tasks seem to correspond to exploration of the space (green border), while others are more directed towards specific areas (blue border). Comparing tasks earlier and later in the curriculum (step 1 to step 5), we find an increase in structure. 
    }\label{fig:vizdoom_tasks}
\end{figure}

\textbf{Visualizing the task distribution}. Modeling pixel observations reveals trajectory-level organization in the underlying true state space (Figure \ref{fig:vizdoom_tasks}). Each map portrays trajectories of a mixture component, with position encoded in 2D space and orientation encoded in the jet color-space; an example of interpreting the maps is shown left of the legend. The components of the mixture model reveal structured groups of trajectories: some components correspond to exploration of the space (marked with green border), while others are more strongly directed towards specific areas (blue border).
The skill maps of the fixed and random environments are qualitatively different: tasks in the fixed room tend towards interactions with objects or walls, while many of the tasks in the random setting sweep the space in a particular direction.
We can also see the evolution of the task distribution at earlier and later stages of Algorithm~\ref{alg:ours}. While initial tasks (produced by a randomly initialized policy) tend to be less structured, we later see refinement of certain tasks as well as the emergence of others as the agent collects new data and acquires strategies for performing existing tasks.

\vspace{-1pt}
\textbf{Do acquired skills transfer to test tasks}? We evaluate how well the CARML task distribution prepares the agent for unseen tasks. For both the fixed and randomized CMP experiments, each test task specifies a dense goal-distance reward for reaching a single object in the environment. In the randomized environment setting, the target objects at test-time are held out from meta-training. The PPO and RND-initialized baseline polices, and the finetuned \method{} meta-policy, are trained for a single target (a specific object in a fixed environment), with 100 episodes per PPO policy update.

\begin{figure}[!t]
    \centering
    \begin{subfigure}[c]{0.32\textwidth}
      \centering\includegraphics[width=\textwidth]{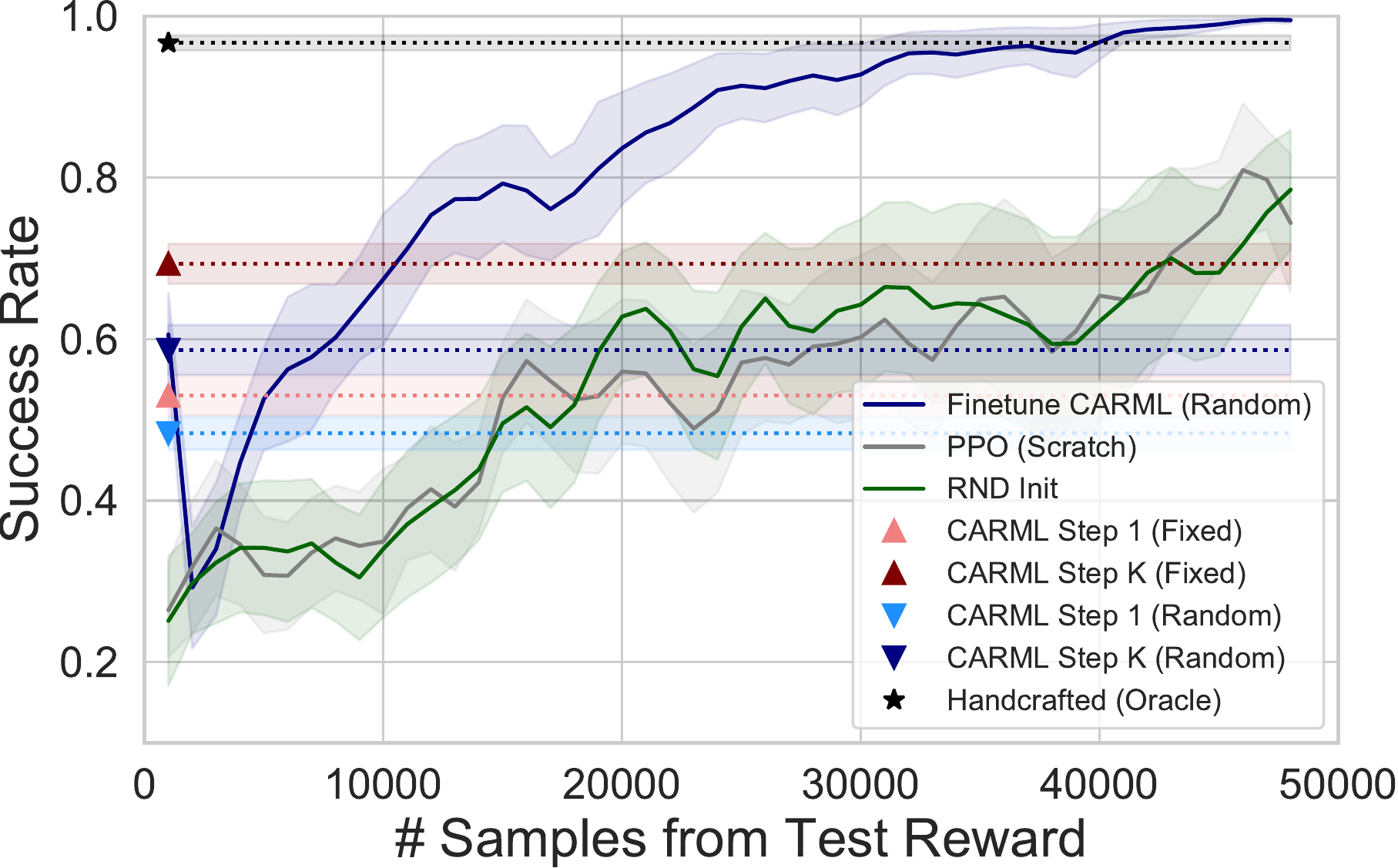}
      \vspace{-0.4cm}
      \caption{\footnotesize ViZDoom}
      \label{fig:vizdoom_fixed_transfer}
    \end{subfigure}
    \begin{subfigure}[c]{0.3\textwidth}
      \centering\includegraphics[width=\textwidth]{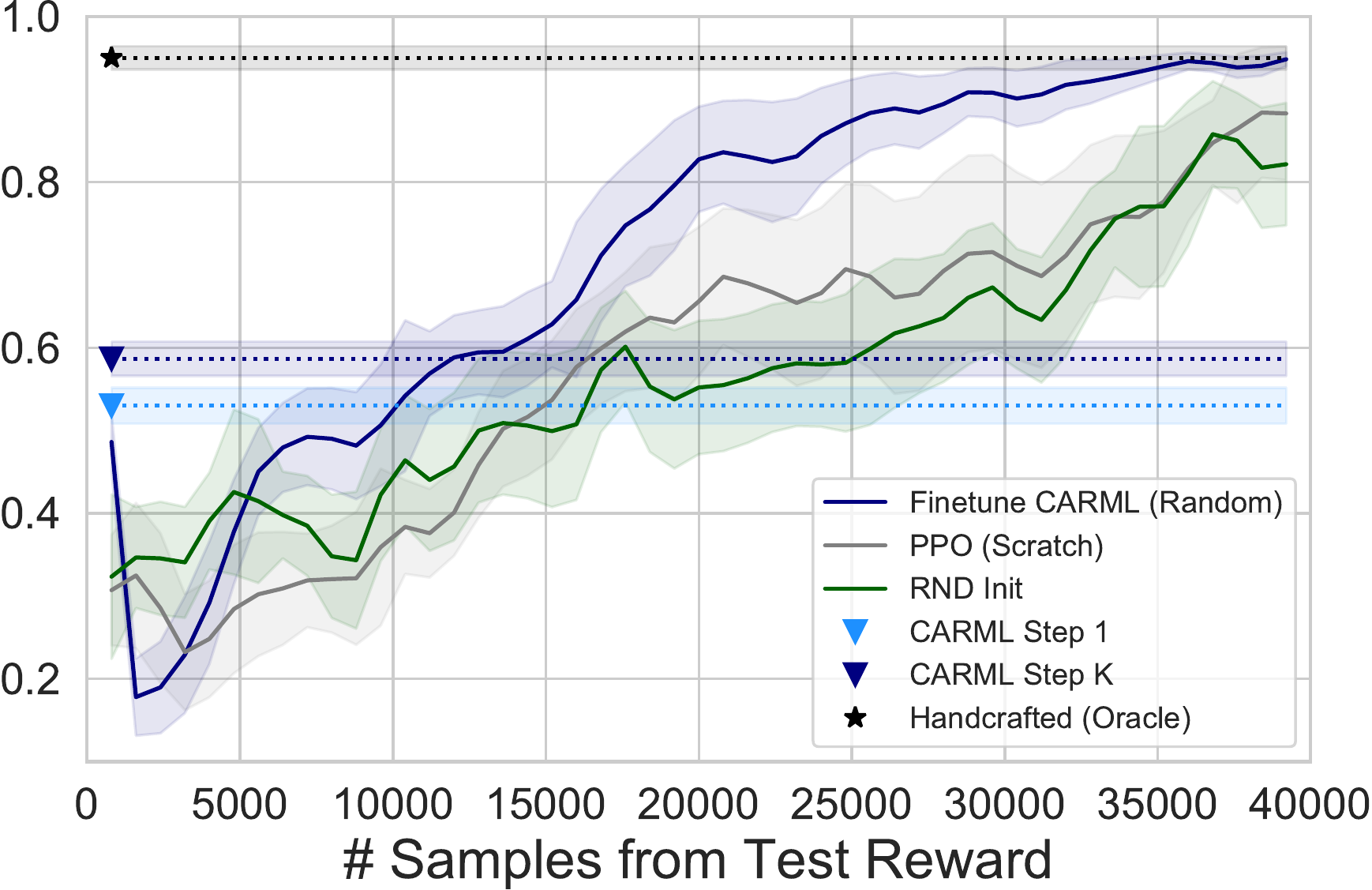}
      \vspace{-0.4cm}
      \caption{\footnotesize Sawyer}
      \label{fig:sawyer_transfer}
    \end{subfigure}
    \begin{subfigure}[c]{0.3\textwidth}
      \centering\includegraphics[width=\textwidth]{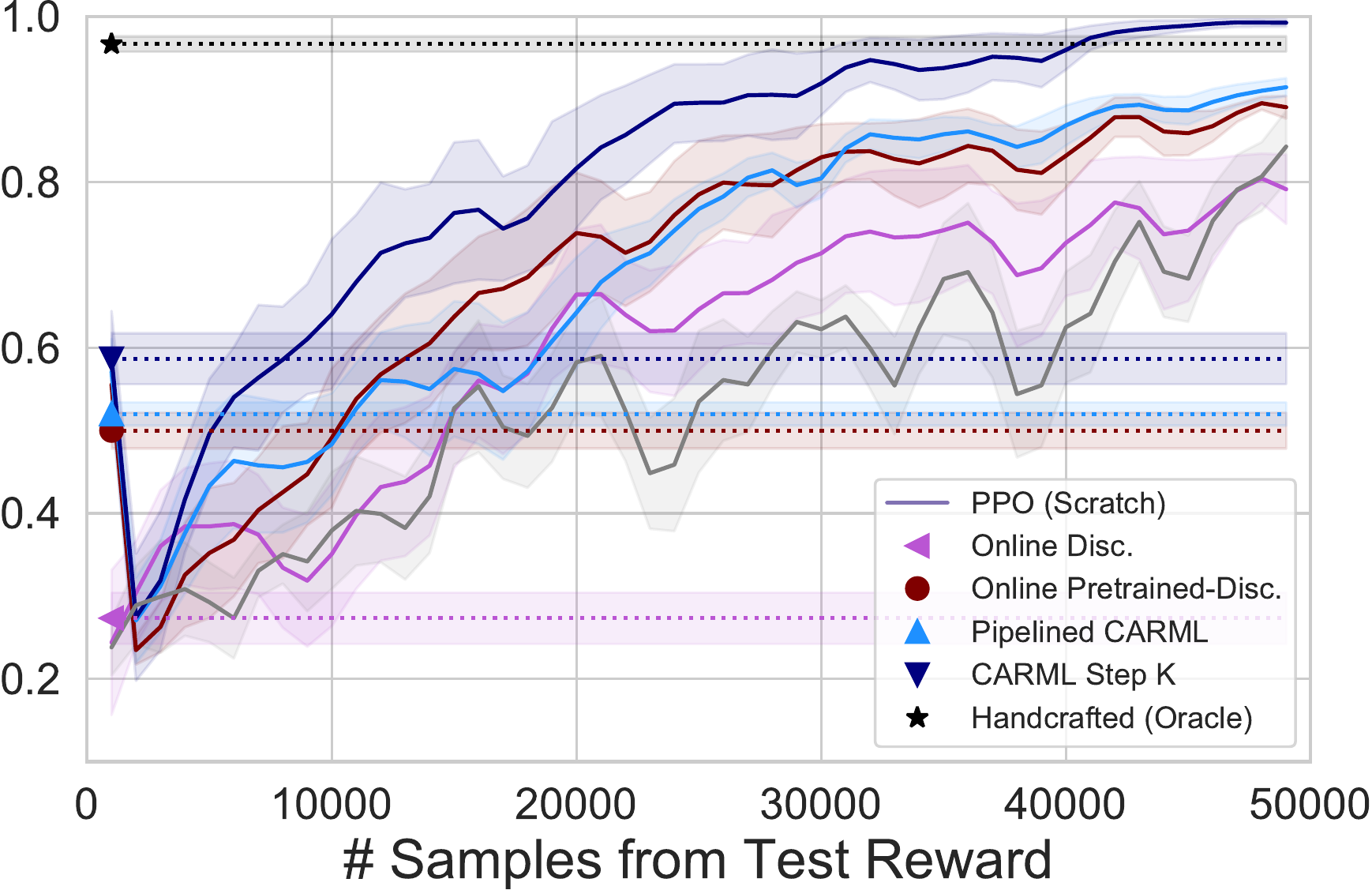}
      \vspace{-0.4cm}
      \caption{\footnotesize Variants (ViZDoom Random)}
      \label{fig:vizdoom_variants}
    \end{subfigure}
    \vspace{-0.5em}
    
    \caption{\footnotesize \method{} enables unsupervised meta-learning of skills that transfer to downstream tasks. Direct transfer curves (marker and dotted line) represent a meta-learner deploying for just 200 time steps at test time. Compared to \method{}, PPO and RND Init sample the test reward function orders of magnitude more times to perform similarly on a single task. Finetuning the \method{} policy also allows for solving individual tasks with significantly fewer samples. The ablation experiments (c) assess both direct transfer and finetuning for each variant. Compared to variants, the \method{} task acquisition procedure results in improved transfer due to mitigation of task mode-collapse and adaptation of the task distribution.}
    \label{fig:transfer}
    \vspace{-7mm}
\end{figure}

In Figure~\ref{fig:vizdoom_fixed_transfer}, we compare the success rates on test tasks as a function of the number of samples with supervised rewards seen from the environment. Direct transfer performance of meta-learners is shown as points, since in this setting the RL$^2$ agent sees only \emph{four episodes} (200 samples) at test-time, without any parameter updates. 
We see that direct transfer is significant, achieving up to 71\%  and 59\% success rates on the fixed and randomized settings, respectively. The baselines require over two orders of magnitude more test-time samples to solve a single task at the same level. 

While the \method{} meta-policy does not consistently solve the test tasks, this is not surprising since no information is assumed about target reward functions during unsupervised meta-learning; inevitable discrepancies between the meta-train and test task distributions will mean that meta-learned strategies \textit{will} be suboptimal for the test tasks. For instance, during testing, the agent sometimes `stalls' before the target object (once inferred), in order to exploit the inverse distance reward. Nevertheless, we also see that finetuning the \method{} meta-policy \textit{trained on random} environments on individual tasks is more sample efficient than learning from scratch. This suggests that deriving reward functions from our mixture model yields useful tasks insofar as they facilitate learning of strategies that transfer.

\vspace{-1pt}
\textbf{Benefit of reorganization}. In Figure~\ref{fig:vizdoom_fixed_transfer}, we also compare performance across early and late outer-loop iterations of Algorithm \ref{alg:ours}, to study the effect of adapting the task distribution (the \method{} E-step) by reorganizing tasks and incorporating new data. In both cases, number of outer-loop iterations $K=5$.
Overall, the refinement of the task distribution, which we saw in Figure~\ref{fig:vizdoom_tasks}, leads improved to transfer performance. 
The effect of reorganization is further visualized in the Appendix~\ref{app:sec:task_distribution_evolution}.

\vspace{-1  pt}
\textbf{Variants}. From Figure~\ref{fig:vizdoom_variants}, we see that the purely online discriminator variant suffers in direct transfer performance; this is due to the issue of mode-collapse in task distribution, wherein the task distribution lacks diversity. Pretraining the discriminator encoder with Algorithm~\ref{alg:em} mitigates mode-collapse to an extent, improving task diversity as the features and task decision boundaries are first fit on a corpus of (randomly collected) trajectories. Finally, while the distribution of tasks eventually discovered by the pipelined variant may be diverse and structured,
meta-learning the corresponding tasks from scratch is harder. More detailed analysis and visualization is given in Appendix~\ref{app:sec:variants}.

\vspace{-0.27cm}
\subsection{Visual Robotic Manipulation}
\vspace{-0.27cm}

\begin{figure*}[t!]
    \centering
        \begin{subfigure}[c]{1\linewidth}
          \centering\includegraphics[width=\textwidth]{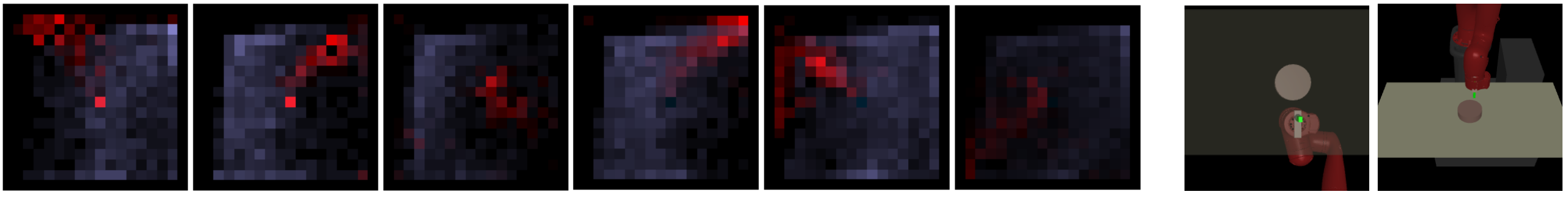}
        \end{subfigure}\hspace{2mm}
        \vspace{-0.2cm}
    \caption{(Left) Skill maps for visuomotor control. Red encodes the true position of the object, and light blue that of the end-effector. Tasks correspond to moving the object to various regions (see Appendix~\ref{app:sec:sawyer} for more skills maps and analysis). (Right) Observation and third person view from the environment, respectively.}
    \label{fig:sawyer_tasks}
    \vspace{-4mm}
\end{figure*}

To experiment in a domain with different challenges, we consider a simulated Sawyer arm interacting with an object in MuJoCo~\citep{todorov2012mujoco}, with end-effector continous control in the 2D plane.  The observation is a bottom-up view of a surface supporting an object (Figure \ref{fig:sawyer_tasks}); the camera is stationary, but the view is no longer egocentric and part of the observation is proprioceptive. The test tasks involve pushing the object to a goal (drawn from the set of reachable states), where the reward function is the negative distance to the goal state. A subset of the skill maps is provided below.

\textbf{Do acquired skills directly transfer to test tasks}? In Figure \ref{fig:sawyer_transfer}, we evaluate the meta-policy on the test task distribution, comparing against baselines as previously. Despite the increased difficulty of control, our approach allows for meta-learning skills that transfer to the goal distance reward task distribution. We find that transfer is weaker compared to the visual navigation (fixed version): one reason may be that the environment is not as visually rich, resulting in a significant gap between the CARML and the object-centric test task distributions.

\begin{wrapfigure}{r}{0.53 \textwidth}
    \vspace{-10mm}
    \centering
    \begin{subfigure}[t]{0.50\linewidth}
      \centering\includegraphics[width=\textwidth]{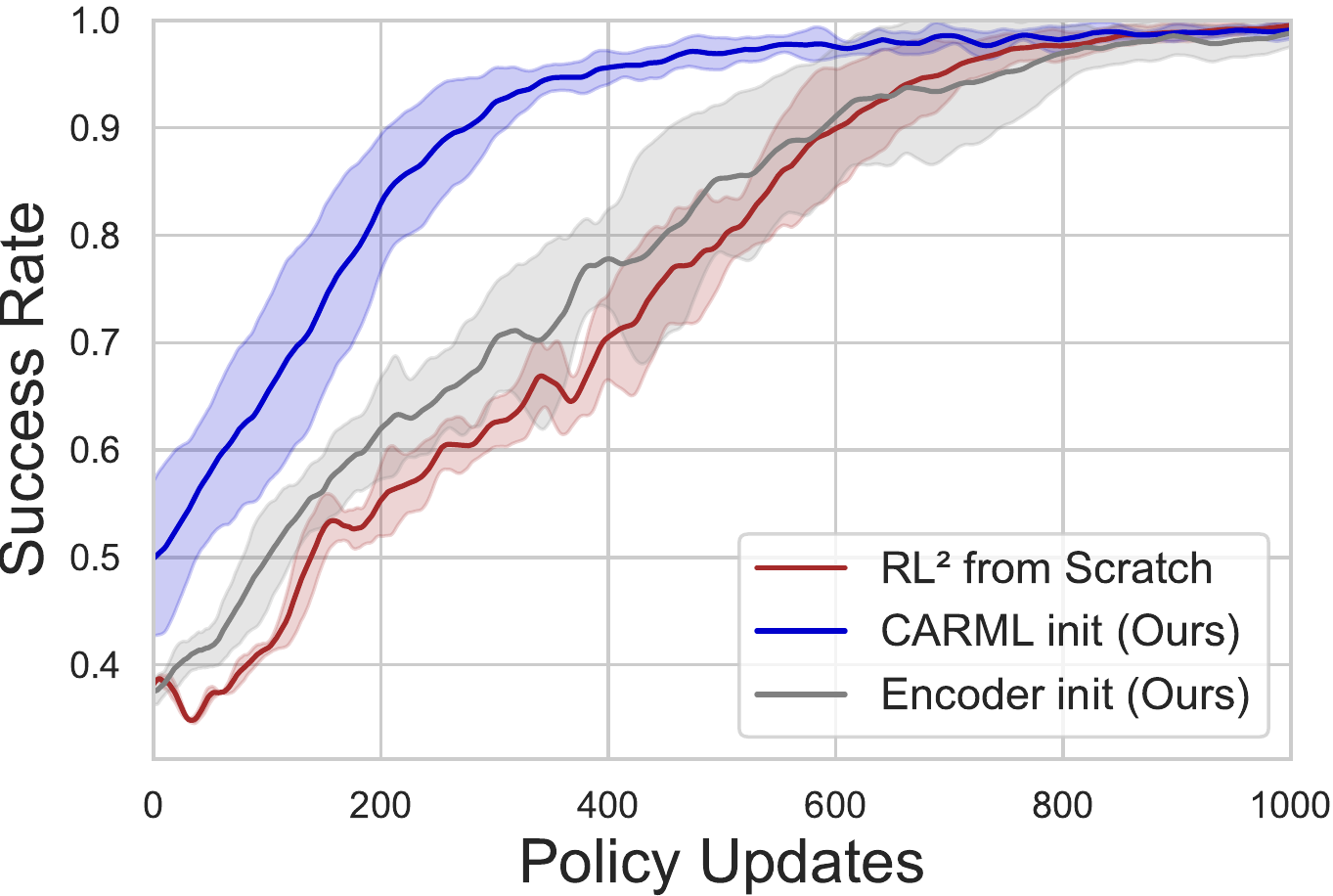}
      \vspace{-4mm}
      \caption{\scriptsize ViZDoom (random)}
        \label{fig:vizdoom_finetune}
    \end{subfigure}
    \begin{subfigure}[t]{0.475\linewidth}
      \centering\includegraphics[width=\textwidth]{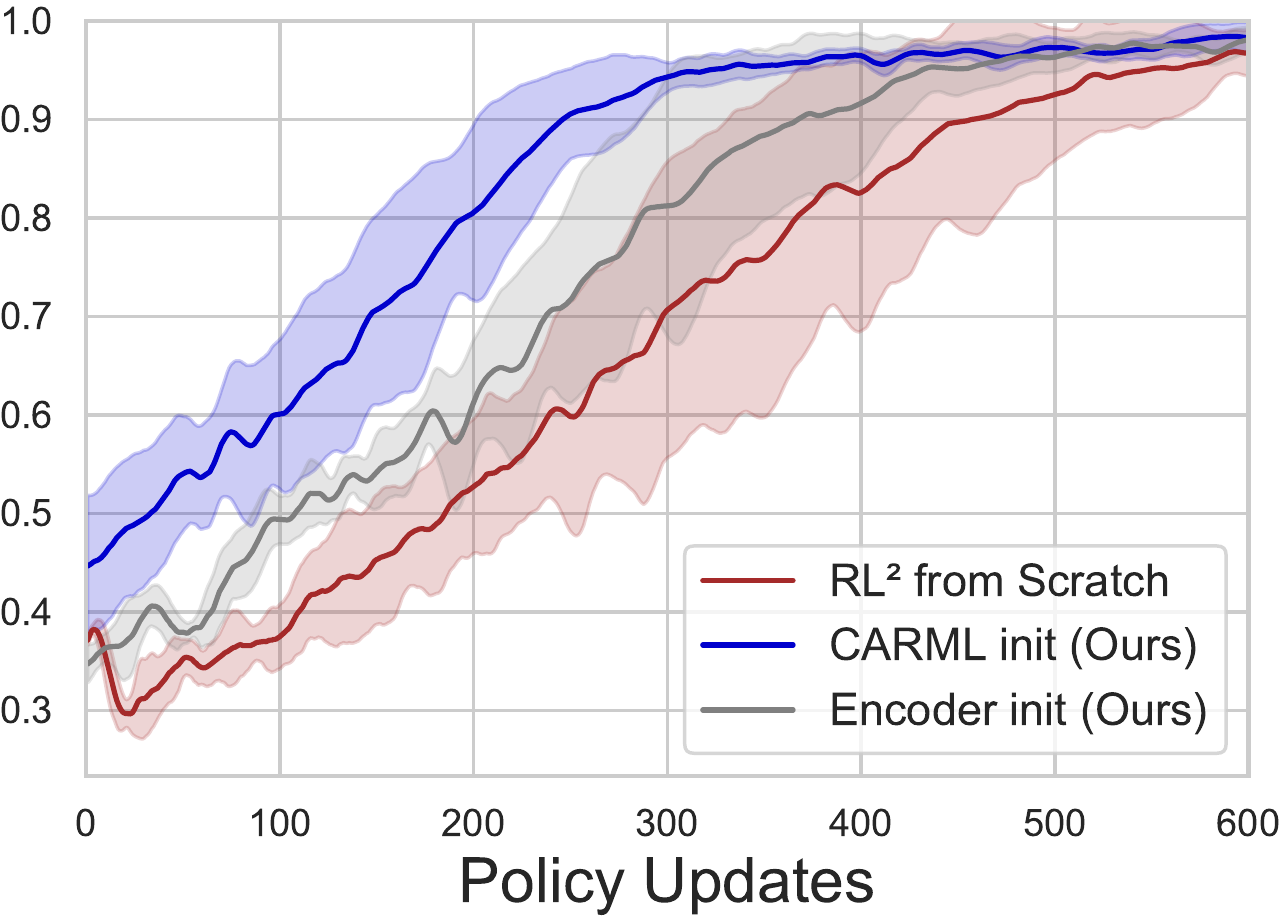}
      \vspace{-4mm}
      \hspace{-3mm}\caption{\scriptsize Sawyer}
        \label{fig:sawyer_finetune}
    \end{subfigure}
    \vspace{-1mm}
    \caption{\!Finetuning the CARML meta-policy allows for accelerated meta-learning of the target task distribution. Curves reflect error bars across three random seeds. \label{fig:finetune}}
    \vspace{-0.40cm}
\end{wrapfigure}

\vspace{-0.3cm}
\subsection{CARML as Meta-Pretraining}
\vspace{-0.3cm}
Another compelling form of transfer is pretraining of an initialization for accelerated supervised meta-RL of target task distributions. 
In Figure~\ref{fig:finetune}, we see that the initialization learned by CARML enables effective supervised meta-RL with significantly fewer samples. To separate the effect of the learning the recurrent meta-policy and the visual representation, we also compare to only initializing the pre-trained encoder.
Thus, while direct transfer of the meta-policy may not directly result in optimal behavior on test tasks, accelerated learning of the test task distribution suggests that the acquired meta-learning strategies may be useful for learning related task distributions, effectively acting as pre-training procedure for meta-RL.

\vspace{-3mm}
\section{Discussion}
\vspace{-3mm}
We proposed a framework for inducing unsupervised, adaptive task distributions for meta-RL that scales to environments with high-dimensional pixel observations. Through experiments in visual navigation and manipulation domains, we showed that this procedure enables unsupervised acquisition of meta-learning strategies that transfer to downstream test task distributions in terms of direct evaluation, more sample-efficient fine-tuning, and more sample-efficient supervised meta-learning. Nevertheless, the following key issues are important to explore in future work.

\vspace{-2pt}
\textbf{Task distribution mismatch}. While our results show that useful structure can be meta-learned in an unsupervised manner, results like the stalling behavior in ViZDoom (see \cref{sec:viznav}) suggest that direct transfer of unsupervised meta-learning strategies suffers from a no-free-lunch issue: there will always be a gap between unsupervised and downstream task distributions, and more so with more complex environments. Moreover, the semantics of target tasks may not necessarily align with especially discriminative visual features. This is part of the reason why transfer in the Sawyer domain is less successful. Capturing other forms of structure useful for  stimulus-reward association might involve incorporating domain-specific inductive biases into the task-scaffold model.
Another way forward is the semi-supervised setting, whereby data-driven bias is incorporated at meta-training time.

\vspace{-3pt}
\textbf{Validation and early stopping}: Since the objective optimized by the proposed method is non-stationary and in no way guaranteed to be correlated with objectives of test tasks, one must provide some mechanism for validation of iterates.

\vspace{-3pt}
\textbf{Form of skill-set}. For the main experiments, we fixed a number of discrete tasks to be learned (without tuning this), but one should consider how the set of skills can be grown or parameterized to have higher capacity (e.g. a multi-label or continuous latent). Otherwise, the task distribution may become overloaded (complicating task inference) or limited in capacity (preventing coverage).

\vspace{-3pt}
\textbf{Accumulation of skill}. We mitigate forgetting with the simple solution of reservoir sampling. Better solutions involve studying an intersection of continual learning and meta-learning.

\vspace{-3pt}

\newpage
\subsubsection*{Acknowledgments} We thank the BAIR community for helpful discussion, and Michael Janner and Oleh Rybkin in particular for feedback on an earlier draft. AJ thanks Alexei Efros for his steadfastness and advice, and Sasha Sax and Ashish Kumar for discussion. KH thanks his family for their support. AJ is supported by the PD Soros Fellowship. This work was supported in part by the National Science Foundation, IIS-1651843, IIS-1700697, and IIS-1700696, as well as Google.

\bibliography{neurips_2019}
\bibliographystyle{plainnat}

\newpage

\begin{appendices}

\section{Derivations}
\subsection{Derivation for Trajectory-Level Responsibilities (Section 3.2.1)}\label{app:traj_derivation}
Here we show that, assuming independence between states in a trajectory when conditioning on a latent variable, computing the trajectory likelihood as a factorized product of state likelihoods for the E-step in standard EM forces the component responsibilities for all states in the trajectory to be identical. Begin by lower-bounding the log-likelihood of the trajectory dataset with Jensen's inequality:
\begin{align}
    \sum_i \log p(\traj) = &\sum_i \log p(\state_1^i, \state_2^i, ..., \state_T^i)\\
    = &\sum_i \log \sum_z p(\state_1^i, \state_2^i, ..., \state_T^i | z) p(z)\\
    \geq &\sum_i \sum_z q_\phi(z | \state_1, \state_2, ..., \state_T) \log \frac{p(\state_1^i, \state_2^i, ..., \state_T^i | z) p(z)}{q_\phi(z | \state_1, \state_2 ...\state_T)}\\
    = 
    &\sum_i \mathbb{E}_{z \sim q_\phi(z | \state_1, \state_2, ..., \state_T)} \log \frac{p(\state_1^i, \state_2^i, ..., \state_T^i | z) p(z)}{q_\phi(z | \state_1, \state_2, ..., \state_T)}. \label{eq:objective}
\end{align}
We have introduced the variational distribution $q_\phi(\traj, z)$, where $z$ is a categorical variable.
Now, to maximize Eq.~\ref{eq:objective} with respect to $\phi := (\bm{\mu}_1, \bm{\Sigma}_1, \pi_1, ..., \bm{\mu}_N, \bm{\Sigma}_N, \pi_N) $, we alternate between an E-step and an M-step, where the E-step is computing
\begin{align}
    q_{ik} &= q_\phi(z = k | \state_1^i, \state_2^i, ..., \state_T^i)\\
    &= \frac{q_\phi(\state_1^i, \state_2^i, ..., \state_T^i | z=k)q_\phi(z=k)}{\sum_j q_\phi(\state_1^i, \state_2^i, ..., \state_T^i | z=j)q_\phi(z=j)}\\
    &= \frac{q_\phi(\state_1^i | z=k)q_\phi(\state_2^i | z=k) \cdots q_\phi(\state_T^i | z=k)q_\phi(z=k)}{\sum_j q_\phi(\state_1^i | z=j)q_\phi(\state_2^i | z=j) \cdots q_\phi(\state_T^i | z=j)q_\phi(z=j)}.
\end{align}

We assume that each $q_\phi(\state | z=k)$ is Gaussian; the M-step amounts to computing the maximum-likelihood estimate of $\phi$, under the mixture responsibilities from the E-step:
\begin{align}
    &\bm{\mu}_k = \frac{\sum_{i=1}^N \frac{q_{ik}}{T}\sum_{t=1}^T \state_t}{\sum_{i=1}^N q_{ik}}\\
    &\bm{\Sigma}_k = \frac{\sum_{i=1}^N \frac{q_{ik}}{T}\sum_{t=1}^T (\state_t - \bm{\mu}_k)(\state_t - \bm{\mu}_k)^\top}{\sum_{i=1}^N q_{ik}}\\
    &\pi_k = \frac{1}{N}\sum_{i=1}^N q_{ik}.
\end{align}

In particular, note that the expressions are independent of $t$. Thus, the posterior $q_\phi(z|\state)$ will be, too.

\subsection{CARML M-Step}
The objective used to optimize the meta-RL algorithm in the CARML M-step can be interpreted as a sum of cross entropies, resulting in the mutual information plus two additional KL terms:

\begin{align} 
&-\mathbb{E}_{\state\sim \pi_\theta(\state|\z), \z \sim q_\phi(\z)} \big[ \log q_\phi(\state) -  \log q_\phi(\state|\z) \big] \\
=& -\sum_\z q_\phi(\z) \sum_\state \pi_\theta(\state|\z) \left(\log q_\phi(\state) - \log q_\phi(\state|\z)\right) \\
=& - \sum_\z q_\phi(\z) \sum_\state \pi(\state|\z) \left(\log \frac{q_\phi(\state)}{\pi_\theta(\state)} + \log \pi_\theta(\state) - \log \frac{q_\phi(\state|\z)}{\pi_\theta(\state|\z)} - \log \pi_\theta(\state|\z)\right) \\
=& H(\pi_\theta(\state)) + \infdiv{\pi_\theta(\state)}{q_\phi(\state)} - H(\pi_\theta(\state|\z)) - \infdiv{\pi_\theta(\state|\z)}{q_\phi(\state|\z)} \\
=& I(\pi_\theta(\state); q_\phi(\z)) + \infdiv{\pi_\theta(\state)}{q_\phi(\state)} - \infdiv{\pi_\theta(\state|\z)}{q_\phi(\state|\z)}.
\end{align}

The first KL term can be interpreted as encouraging exploration with respect to the density of the mixture. The second KL term is the reverse KL term for matching the modes of the mixture. 

\textbf{Density-based exploration}.
In practice, we may want to trade off between exploration and matching the modes of the generative model:

\begin{align}
    r_\z(\state) &= \lambda \log q_\phi(\state|\z) - \log q_\phi(\state) \label{app:eq:gen_reward} \\
    &= (\lambda - 1)  \log  q_\phi(\state|\z) + \log q_\phi(\z|\state) - \log q_\phi(\z) \label{app:eq:gen_reward2} \\
    &= (\lambda - 1)  \log  q_\phi(\state|\z) + \log q_\phi(\z|\state) + C \label{app:eq:gen_reward3}
\end{align}
where $C$ is constant with respect to the optimization of $\theta$.
Hence, the objective amounts to maximizing discriminability of skills where $\lambda < 1$ yields a bonus for exploring away from the mode of the corresponding skill.

\subsection{Discriminative \method{} and DIAYN} \label{app:sec:diayn}
Here, we derive a discriminative instantiation of \method{}. We begin with the E-step. We leverage the same conditional independence assumption as before, and re-write the trajectory-level MI as the state level MI, assuming that trajectories are all of length $T$:
\begin{equation}
    I(\traj;\z) \geq \frac{1}{T} \sum_t I(\state_t ; \z) = I(\state ; \z)
\end{equation}
We then decompose MI as the difference between marginal and conditional entropy of the latent, and choose the variational distribution to be the product of a classifier $q_{\phi_c}(\z | \state)$ and a density model $q_{\phi_d}(\state)$:
\begin{align}
    I(\state;\z) &= H(\z) - H(\z|\state) \\
    &= -\sum_\z p(\z) \log p(\z) + \sum_{\state,\z} \pi_\theta(\state, \z) \log \pi_\theta(\z|\state) \\
    & \geq -\sum_\z p(\z) \log p(\z) + \sum_{\state,\z} \pi_\theta(\state | \z) p(\z) \log q_{\phi_c}(\z|\state) \label{eq:infomax}
\end{align}
We fix $\z$ to be a uniformly-distributed categorical variable. The \method{} E-step consists of two separate optimizations: supervised learning of $q_{\phi_c}(\z | \state)$ with a cross-entropy loss and density estimation of $q_{\phi_d}(\state)$:
\begin{equation}\label{eq:discriminator}
    \max_{\phi_c} \; \E_{\z \sim p(\z), \state \sim \pi_\theta(\z)} \left[\log q_{\phi_c}(\z | \state) \right] \qquad 
    \max_{\phi_d} \; \E_{\z \sim p(\z), \state \sim \pi_\theta(\z)} \left[\log q_{\phi_d}(\state) \right]
\end{equation}
For the \method{} M-step, we start from the form of the reward in Eq.~\ref{app:eq:gen_reward2} and manipulate via Bayes':
\begin{align}
    r_\z(\state) 
    &= \log  q_{\phi_c}(\z|\state) + (\lambda - 1) \log q_{\phi_c}(\z|\state) + (\lambda - 1) \log q_{\phi_d}(\state) - (\lambda - 1) \log p(\z) - \log p(\z) \nonumber \\
    &= \lambda \log q_{\phi_c}(\z|\state) + (\lambda - 1) \log q_{\phi_d}(\state) + C
\end{align}
where $C$ is constant with respect to the optimization of $\theta$ in the M-step
\begin{equation}
    \max_{\theta} \; \E_{\z \sim q_\phi(\z), \state \sim \pi_\theta(\z)} \big[ 
        \lambda \log q_{\phi_c}(\z|\state) + (\lambda - 1) \log q_{\phi_d}(\state)
    \big] \label{eq:disc_pol}
\end{equation}
To enable a trajectory-level latent $\z$, we want every state in a trajectory to be classified to the same $\z$. This is achievable in a straightforward manner: when training the classifier $q_{\phi_c}(\z | \state)$ via supervised learning, label each state in a trajectory with the realization of $\z$ that the policy $\pi_\theta(\action | \state, \z)$ was conditioned on when generating that trajectory.

\textbf{Connection to DIAYN}. Note that with $\lambda = 1$ in Eq.~\ref{eq:disc_pol}, we directly obtain the DIAYN~\citep{eysenbach2019diversity} objective without standard policy entropy regularization, and we do away with needing to maintain a density model $\log q_{\phi_d}(\state)$, leaving just the discriminator. If $\pi_\theta(\action | \state, \z)$ is truly a contextual policy (rather than the policy given by adapting a meta-learner), we have recovered the DIAYN algorithm. This allows us to interpret on DIAYN-style algorithms as implicitly doing trajectory-level clustering with a conditional independence assumption between states in a trajectory given the latent. This arises from the weak trajectory-level supervision specified when training the discriminator: all states in a trajectory are assumed to correspond to the same realization of the latent variable.

\newpage
\section{Additional Details for Main Experiments} \label{app:sec:implementation_details}
\subsection{CARML Hyperparameters}
We train CARML for five iterations, with 500 PPO updates for meta-learning with RL$^2$ in the M-step (i.e. update the mixture model every 500 meta-policy updates). Thus, the CARML unsupervised learning process consumes on the order of 1,000,000 episodes (compared to the \textasciitilde400,000 episodes needed to train a meta-policy with the true task distribution, as shown in our experiments). We did not heavily tune this number, though we noticed that using too few policy updates (e.g. \textasciitilde100) before refitting $q_\phi$ resulted in instability insofar as the meta-learner does not adapt to learn the updated task distribution. Each PPO learning update involves sampling 100 tasks with 4 episodes each, for a total of 400 episodes per update. We use 10 PPO epochs per update with a batch size of 100 tasks.

During meta-training, tasks are drawn according to $z \sim q_\phi(z)$, the mixture's latent prior distribution. Unless otherwise stated, we use $\lambda=0.99$ for all visual meta-RL experiments. For all experiments unless otherwise mentioned, we fix the number of components in our mixture to be $k=16$. We use a reservoir of $1000$ trajectories.

\textbf{Temporally Smoothed Reward:} At unsupervised meta-training time, we found it helpful to reward the meta-learner with the average over a small temporal window, i.e. $r^W_z(s_t) = \frac{1}{W} \sum_{i=t-W}^t r_z(s_i)$, choosing $W$ to be $W=10$. This has the effect of smoothing the reward function, thereby regularizing acquired task inference strategies.

\textbf{Random Seeds:} The results reported in Figure~6 are averaged across policies (for each treatment) trained with three different random seeds. The performance is averaged across 20 test tasks. The results reported in Figure~7 are based on finetuning CARML policies trained with three different random seeds. We did not observe significant effects of the random seed used in the finetuning procedure of experiments reported for Figure~7.

\textbf{Model Selection:} Models used for transfer experiments are selected by performance on a small held-out validation set (ten tasks) for each task, that does not intersect with the test task.

\subsection{Meta-RL with RL$^2$}
We adopt the recurrent architecture and hyperparameter settings as specified in the visual maze navigation tasks of \citet{duan2016rl}, except we:
\begin{itemize}
    \setlength\itemsep{-0.1em}
    \item Use PPO for policy optimization ($\textrm{clip}=0.2, \textrm{value\_coef}=0.1$)
    \item Set the entropy bonus coefficient $\alpha$ in an environment-specific manner. We use $\alpha = 0.001$ for MuJoCo Sawyer and $\alpha = 0.1$ for ViZDoom.
    \item Enlarge the input observation space to $84 \times 84 \times 3$, adapting the encoder by half the stride in the first convolutional layer.
    \item Increase the size of the recurrent model (hidden state size 512) and the capacity of the output layer of the RNN (MLP with one hidden layer of dimension 256).
    \item Allow for four episodes per task (instead of two), since the tasks we consider involve more challenging task inference.
    \item Use a multi-layer perceptron with one-hidden layer to readout the output for the actor and critic, given the recurrent hidden state.
\end{itemize}

\subsection{Reward Normalization}
A subtle challenge that arises in applying meta-RL across a range of tasks is difference in the statistics of the reward functions encountered, which may affect task inference. 
Without some form of normalization, the statistics of the rewards of unsupervised meta-training tasks versus those of the downstream tasks may be arbitrarily different, which may interfere with inferring the task. This is especially problematic for RL$^2$ (compared to e.g. MAML \cite{finn2017model}), which relies on encoding the reward as a feature at each timestep.
We address this issue by whitening the reward at each timestep with running mean and variance computed online, separately for each task from the unsupervised task distribution during meta-training. At test-time, we share these statistics across tasks from the same test task distribution.

\subsection{Learning Visual Representations with DeepCluster}
To jointly learn visual representations with the mixture model, we adopt the optimization scheme of DeepCluster~\citep{caron2018deep}. The DeepCluster model is parameterized by the weights of a convolutional neural network encoder as well as a $k$-means model in embedding space. It is trained in an EM-like fashion, where the M-step additionally involves training the encoder weights via supervised learning of the image-cluster mapping.

Our contribution is that we employ a modified E-step, as presented in the main text, such that the cluster responsibilities are ensured to be consensual across states in a trajectory in the training data. As shown in our experiments, this allows the model to learn trajectory-level visual representations. 
The full \method{} E-step with DeepCluster is presented below.

\begin{algorithm}[H]
   \caption{\method{} E-Step, a Modified EM Procedure, with DeepCluster}
   \label{app:alg:em}
\begin{algorithmic}[1]
   \STATE {\bfseries Require:} a set of trajectories $\buffer=\{(\state_1,\dots,\state_T)\}_{i=1}^N$
   \STATE Initialize $\phi:=(\phi_w, \phi_m)$, the weights of encoder $g$ and embedding-space mixture model parameters.
   \WHILE{not converged}
        \STATE{Compute $L(\phi_m;\traj, z) = \sum_{\state_t \in \traj} \log q_{\phi_m}(g_{\phi_w}(\state_t) | z)$. \label{app:line:aggregate}} 
        \STATE{Update via MLE: $\phi_m \gets \argmax_{\phi'_m}\sum_{i=1}^N L(\phi'_m;\traj_i,z)$.} 
        \STATE{Obtain training data $\mathcal{D} := \{(\state, y:=\argmax_k q_{\phi_m}(z=k | g_{\phi_w}(\state))\}$.}
        \STATE{Update via supervised learning: $\phi_w \gets \argmax_{\phi'_w} \sum_{(\state, y) \in \mathcal{D}} \log q(y | g_{\phi'_w}(\state))$}.
   \ENDWHILE
   \STATE {\bfseries Return:} a mixture model $q_\phi(\state,z)$
\end{algorithmic}
\end{algorithm}

For updating the encoder weights, we use the default hyperparameter settings as described in \cite{caron2018deep}, except 1) we modify the neural network architecture, using a smaller neural network, ResNet-10~ \cite{kaiming2015} with a fixed number of filters (64) for every convolutional layer, and 2) we use number of components $K=16$, which we did not tune.
We tried using a more expressive Gaussian mixture model with full covariances instead of $k$-means (when training the visual representation), but found that this resulted in overfitting. Hence, we use $k$-means until the last iteration of EM, wherein a Gaussian mixture model is fitted under the resulting visual representation.



\subsection{Environments}

\begin{figure}[!t]
    \centering
    \begin{subfigure}[t]{0.45\linewidth}
        \includegraphics[width=\linewidth]{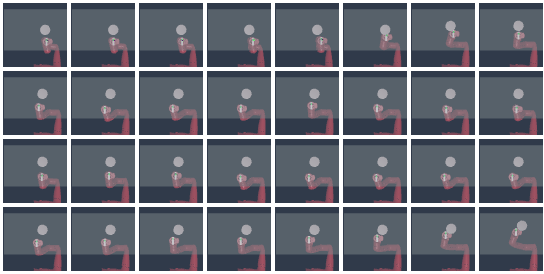}
    \end{subfigure}
    \begin{subfigure}[t]{0.45\linewidth}
        \includegraphics[width=\textwidth]{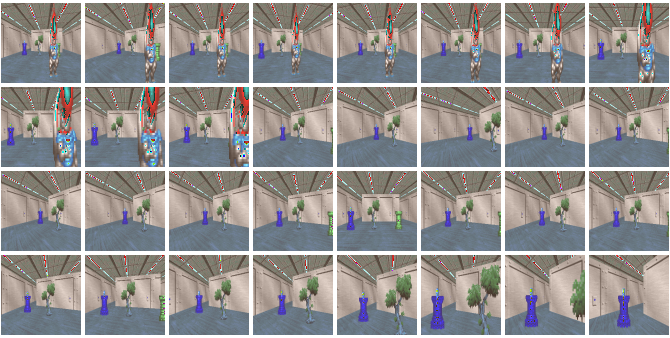}
    \end{subfigure}
    
    \caption{Example Observation Sequences from the Sawyer (left) and Vizdoom Random (right) environments.}
    \label{fig:trajs}
\end{figure}

\subsubsection{ViZDoom Environment} 
\begin{figure}[h]
    \centering
    \includegraphics[width=0.2\textwidth]{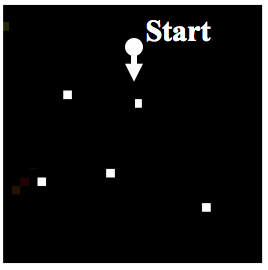}
    \caption{Top-down view of VizDoom environment, with initial agent position. White squares depict stationary objects (only relevant to fixed environment).}
\end{figure}
The environment used for visual navigation is a 500x500 room built with ViZDoom~\citep{kempka2016vizdoom}. We consider both fixed and random environments; for randomly placing objects, the only constraint enforced is that objects should not be within a minimal distance of one another. There are 50 train objects and 50 test objects. The agent's pose is always initialized to be at the top of the room facing forward. We restrict observations from the environment to be $84\times 84$ RGB images. The maximum episode length is set to 50 timesteps. The hand-crafted reward function corresponds to the inverse $l_2$ distance from the specified target object.

The environment considered is relatively simple in layout, but compared to simple mazes, can provide a more complex observation space insofar as objects are constantly viewed from different poses and in various combinations, and are often occluded. The underlying ground-truth state space is the product of continuous 2D position and continuous pose spaces. There are three discrete actions that correspond to turning right, turning left, and moving forward, allowing translation and rotation in the pose space that can vary based on position; the result is that the effective visitable set of poses is not strictly limited to a subset of the pose space, despite discretized actions.

\subsubsection{Sawyer Environment}
\begin{figure}[h]
    \centering
    \includegraphics[width=0.15\textwidth]{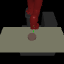}
    \caption{Third person view of the Sawyer environment}
\end{figure}
For visual manipulation, we use a MuJoCo~\citep{todorov2012mujoco} environment involving a simulated Sawyer 7-DOF robotic arm in front of a table, on top of which is an object. The Sawyer arm is controlled by 2D continuous control.
It is almost identical to the environment used by prior work such as \cite{nair2018visual}, with the exception that our goal space is that of the object position. The robot pose and object are always initialized to the same position at the top of the room facing forward. We restrict observations from the environment to be $84\times 84$ RGB images. The maximum episode length is set to 50 timesteps. The hand-crafted reward function corresponds to the negative $l_2$ distance from the specified target object.

\section{Additional Details for Qualitative Study of $\lambda$} \label{app:sec:vae}
\subsection{Instantiating $q_\phi$ as a VAE}
Three factors motivate the use of a variational auto-encoder (VAE) as a generative model for the 2D toy environment. First, a key inductive bias of DeepCluster, namely that randomly initialized convolutional neural networks work surprisingly well, which \citet{caron2018deep} use to motivate its effectiveness in visual domains, does not apply for our 2D state space. Second, components of a standard Gaussian mixture model are inappropriate for modeling trajectories involving turns. Third, using a VAE allows sampling from a continuous latent, potentially affording an unbounded number of skills.

We construct the VAE model in a manner that enables expressive generative densities $p(\state|z)$ while allowing for computation of the policy reward quantities. 
We set the VAE latent to be $(\z, t)$, where $p(\z,t)=p(\z)p(t)=\mathcal{N}(\bm{0}, \bm{I})\frac{1}{T}$. The form of $p(t)$ follows from restricting the policy to sampling trajectories of length $T$.
We factorize the posterior as $q_\phi(\z,t|\state_{t'})=q(\z|\state_{t'})\delta(t-t')$. Keeping with the idea of having a Markovian reward, we construct the VAE's recognition network such that it takes as input individual states after training. To incorporate the constraint that all states in a trajectory are mapped to the same posterior, we adopt a particular training scheme: we pass in entire trajectories $\state_{1:T}$, and specify the posterior parameters as $\mu_{z} = \frac{1}{T} \sum_t g_{\eta}(\state_t)$ and $\sigma^2_{z} = \frac{1}{T} \sum_t g_{\eta}(\state_t)$. 

The ELBO for this model is
\begin{align}
    &\mathbb{E}_{\z,t \sim q_\phi(\z,t |\state_{t'})}\big[\log q_\phi(\state_{t'}|\z, t) \big] - \infdiv{q_\phi(\z,t |\state_{t'})}{p(\z,t)} \\
    =&\mathbb{E}_{\z\sim q_\phi(\z|\state_{t'})} \big[ \log q_\phi(\state_{t'}|\z, t') \big] - \infdiv{q_\phi(\z|\state_{t'})}{p(\z)} - C\label{eq:elbo}
\end{align}
where $C$ is constant with respect to the learnable parameters. The simplification directly follows from the form of the posterior; we have essentially passed $t'$ through the network unchanged. Notice that the computation of the ELBO for a trajectory leverages the conditional independence in our graphical model.

\subsection{\method{} Details}
Since we are not interested in meta-transfer for this experiment, we simplify the learning problem to training a contextual policy $\pi_\theta(\action | \state, z)$.
To reward the policy using the VAE $q_\phi$, we compute 
\begin{equation}
    r_z(\state) = \lambda \log q_\phi(\state|z) - \log q_\phi(\state)
\end{equation}
where 
\begin{align}
    \log q_\phi(\state|z) = \log \sum_t q_\phi(\state|z,t)p(t) = \log \frac{1}{T} \sum_t q_\phi(\state|z,t)
\end{align}
and we approximate $\log q_\phi(\state)$ by its ELBO (Eq.~\ref{eq:elbo}), substituting the above expression for the reconstruction term.

\section{Sawyer Task Distribution} \label{app:sec:sawyer}
Visualizing the components of the acquired task distribution for the Sawyer domain reveals structure and diversity related to the position of the object as well as the control path taken to effect movement. Red encodes the true position of the object, and light blue that of the end-effector. We find tasks corresponding to moving the object to various locations in the environment, as well as tasks that correspond to moving the arm in a certain way without object interaction.  The tasks provide a scaffold for learning to move the object to various regions of the reachable state space.

Since the Sawyer domain is less visually rich than the VizDoom domain, there may be less visually discriminative states that align with semantics of test task distributions. Moreover, since a large part of the observation is proprioceptive, the discriminative clustering representation used for density modeling captures various  proprioceptive features that may not involve object interaction. The consequences are two-fold: 1) the gap in the CARML and the object-centric test task distributions may be large, and 2) the CARML tasks may be too diverse in-so-far as tasks share less structure, and inferring each task involves a different control problem.

\begin{figure}[H]
    \centering
    \includegraphics[width=\linewidth]{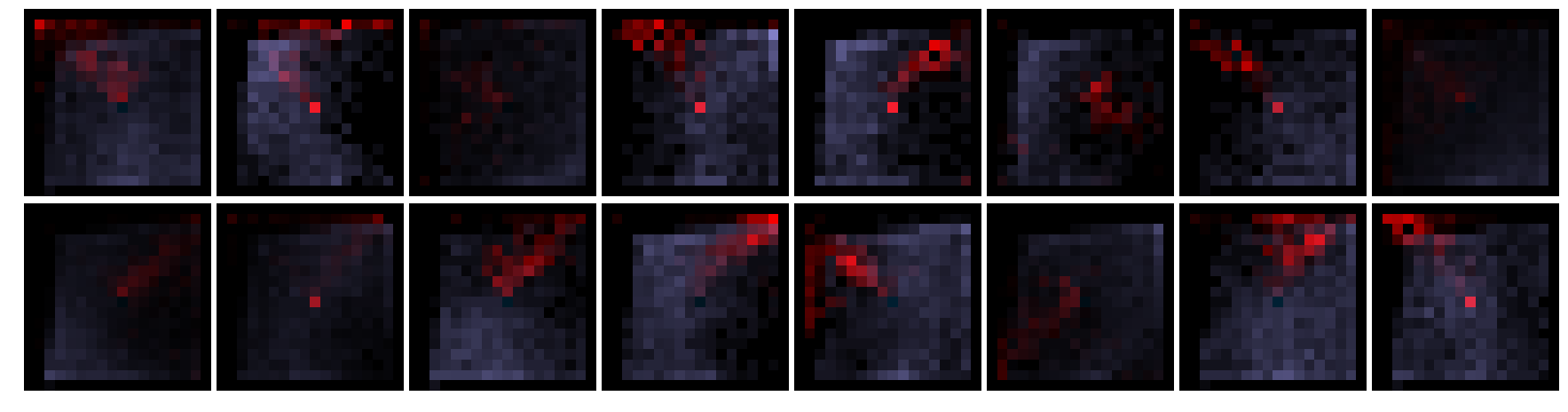}
    \caption{ Skill Maps for Visuomotor Control. Red encodes the true position of the object, and light blue that of the end-effector. The tasks provide a scaffold for learning to move the object to various regions of the reachable state space.}
\end{figure}

\newpage
\section{Mode Collapse in the Task Distribution} \label{app:sec:variants}
Here, we present visualizations of the task distributions induced by variants of the presented method, to illustrate the issue of using an entirely discrimination-based task acquisition approach. Using the fixed VizDoom setting, we compare:

\begin{enumerate}[label=(\roman*)]
  \item CARML, the proposed method
  \item \textbf{online discriminator} -- task acquisition with a purely discriminative $q_\phi$ (akin to an online, pixel-observation-based adaptation of \cite{gupta2018unsupervised});
  \item \textbf{online pretrained-discriminator} -- task acquisition with a discriminative $q_\phi$ as in \textbf{(ii)}, initialized with pre-trained observation encoder.
\end{enumerate}

For all discriminative variants, we found it crucial to use a temperature $\geq 3$ to soften the classifier softmax to prevent immediate task mode-collapse.




\begin{figure}[H]
    \centering
    \captionsetup{justification=centering}
    \begin{subfigure}[t]{0.3\linewidth}
        \includegraphics[width=\linewidth]{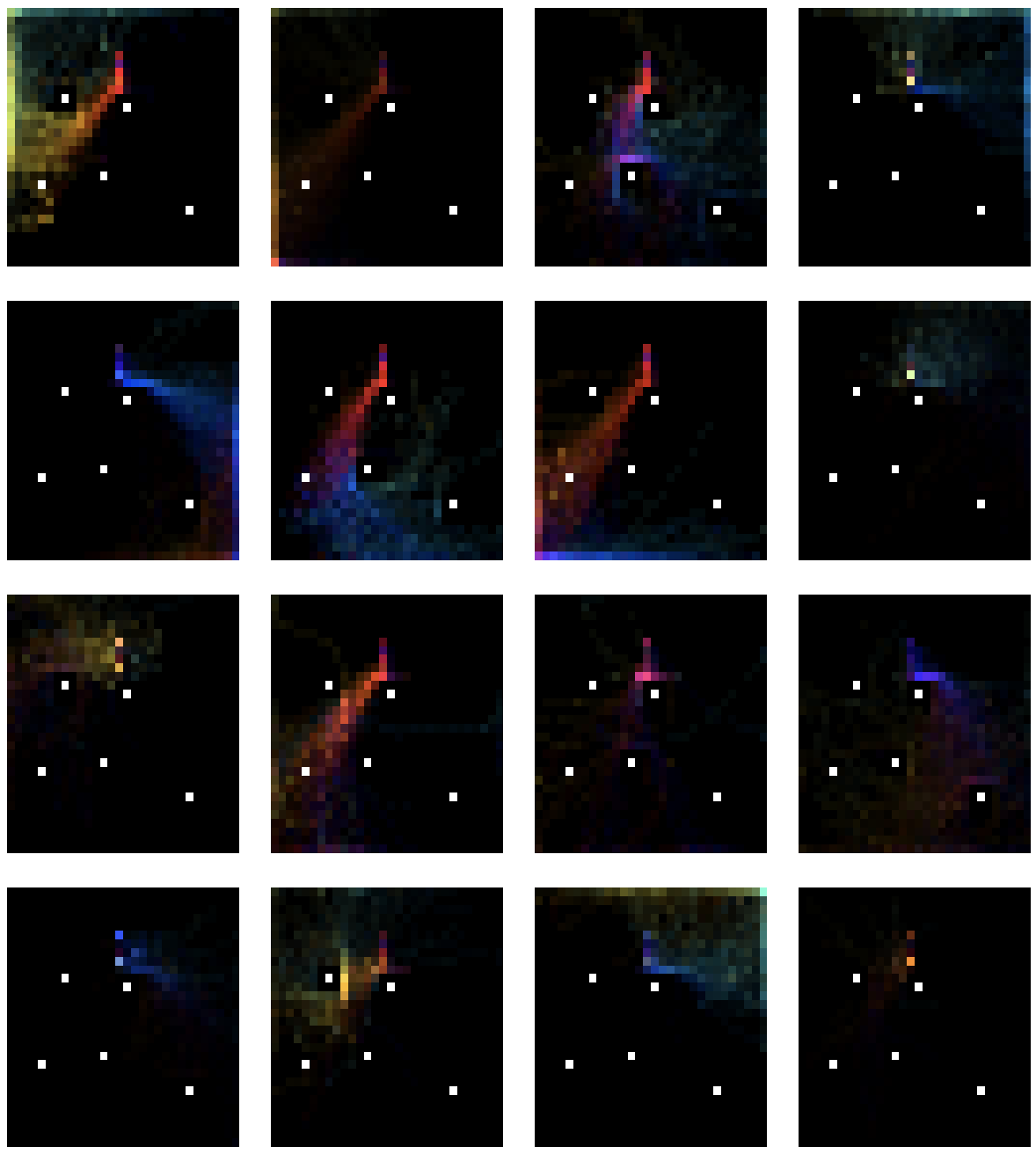}
        \vspace{-15pt}
        \caption*{(i) CARML (ours)}
    \end{subfigure} \hspace{2mm}
    \begin{subfigure}[t]{0.3\linewidth}
        \includegraphics[width=\linewidth]{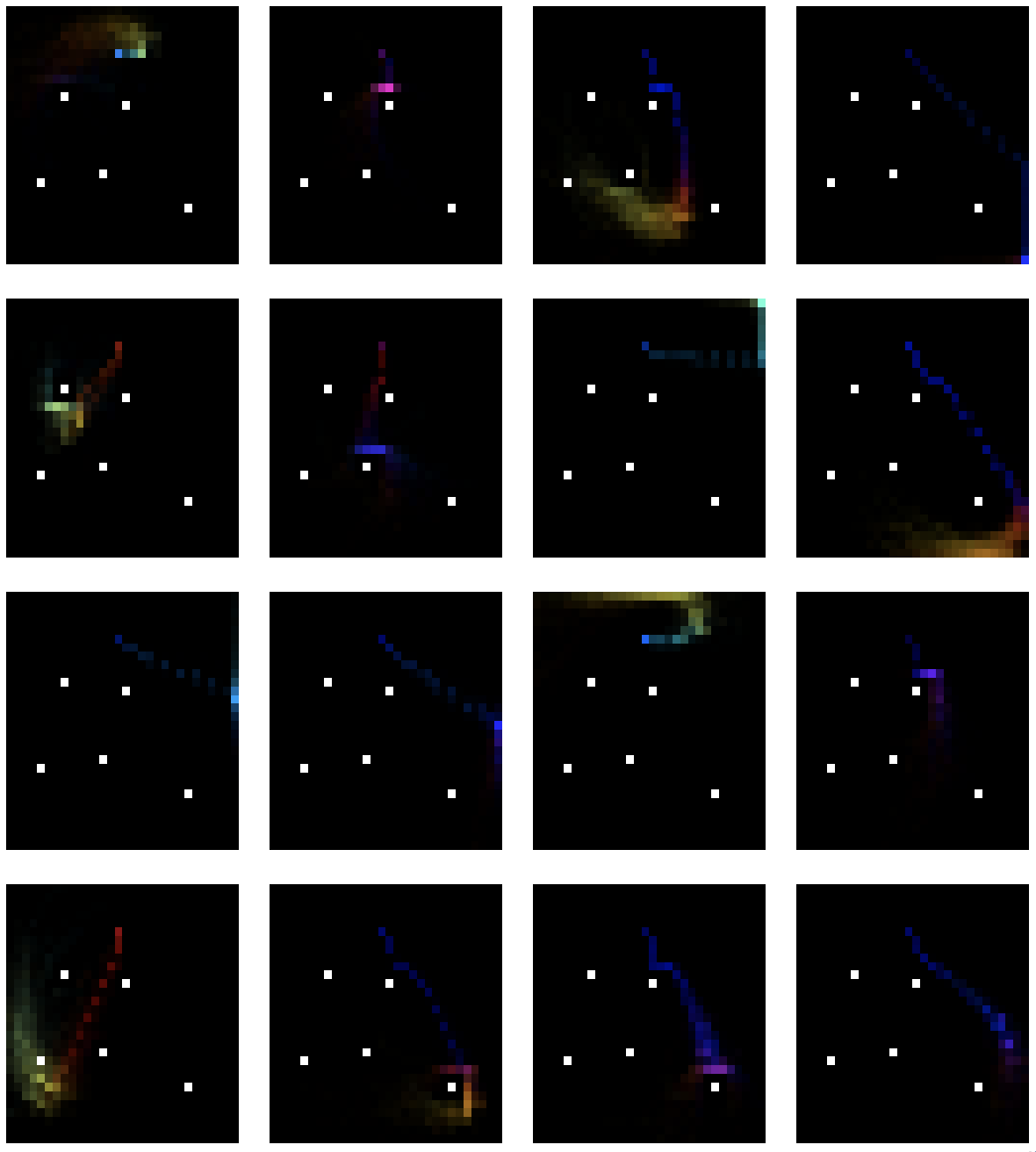}
        \vspace{-15pt}
        \caption*{{(ii) online discriminator \cite{gupta2018unsupervised}}}
    \end{subfigure} \hspace{2mm}
    \begin{subfigure}[t]{0.3\linewidth}
        \includegraphics[width=\linewidth]{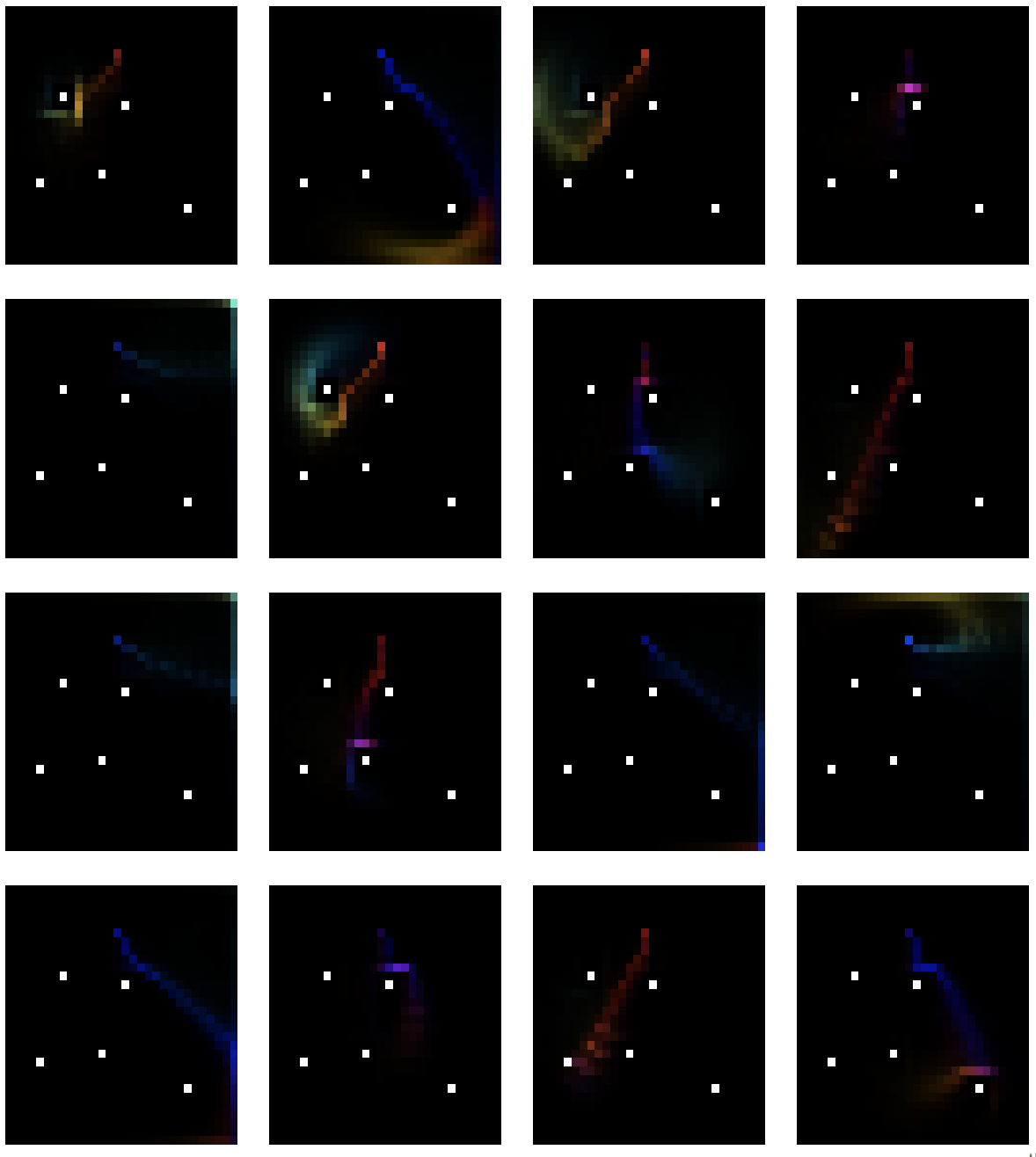}
        \vspace{-15pt}
        \caption*{(iii) pretrained online discriminator}
    \end{subfigure}
    \label{fig:vizdoom_shots}
    \caption*{}
        \vspace{-10mm}
\end{figure}

We find the task acquisition of purely discriminative variants \textbf{(ii, iii)} to suffer from an effect akin to mode-collapse; the policy’s data distribution collapses to a smaller subset of the trajectory space (one or two modes), and tasks correspond to minor variations of these modes. Skill acquisition methods such as DIAYN rely purely on discriminability of states/trajectories under skills, which can be more easily satisfied in high-dimensional observation spaces and can thus lead to such mode-collapse. Moreover, they do not a provide a direct mechanism for furthering exploration once skills are discriminable.

On the other hand, the proposed task acquisition approach (Algorithm~\ref{alg:em}, \cref{sec:e-step}) fits a generative model over jointly learned discriminative features, and is thus not only less susceptible to mode-collapse (w.r.t the policy data distribution), but also allows for density-based exploration (\cref{sec:m-step}). Indeed, we find that \textbf{(iii)} seems to mitigate mode-collapse -- benefiting from a pretrained encoder from \textbf{(i)} -- but does not entirely prevent it. As shown in the main text (Figure~\ref{fig:vizdoom_variants}), in terms of meta-transfer to hand-crafted test tasks, the online discriminative variants \textbf{(ii, iii)} perform worse than CARML \textbf{(i)}, due to lesser diversity in the task distribution. 


\newpage
\section{Evolution of Task Distribution} \label{app:sec:task_distribution_evolution}
Here we consider the evolution of the task distribution in the Random VizDoom environment. The initial tasks (referred to as CARML It. 1) are produced by fitting our deep mixture model to data from a randomly-initialized meta-policy. CARML Its. 2 and 3 correspond to the task distribution after the first and second CARML E-steps, respectively. 

We see that the initial tasks tend to be less structured, in so far as the components appear to be noisier and less distinct. With each E-step we see refinement of certain tasks as well as the emergence of others, as the agent's data distribution is shifted by 1) learning the learnable tasks in the current data-distribution, and 2) exploration. In particular, tasks that are "refined" tend to correspond to more simple, exploitative behaviors (i.e. directly heading to an object or a region in the environment, trajectories that are more straight), which may not require exploration to discover. On the other hand, the emergent tasks seem to reflect exploration strategies (i.e. sweeping the space in an efficient manner). We also see the benefit of reorganization that comes from refitting the mixture model, as tasks that were once separate can be combined.

\begin{figure}[H]
    \centering
    \includegraphics[width=\linewidth]{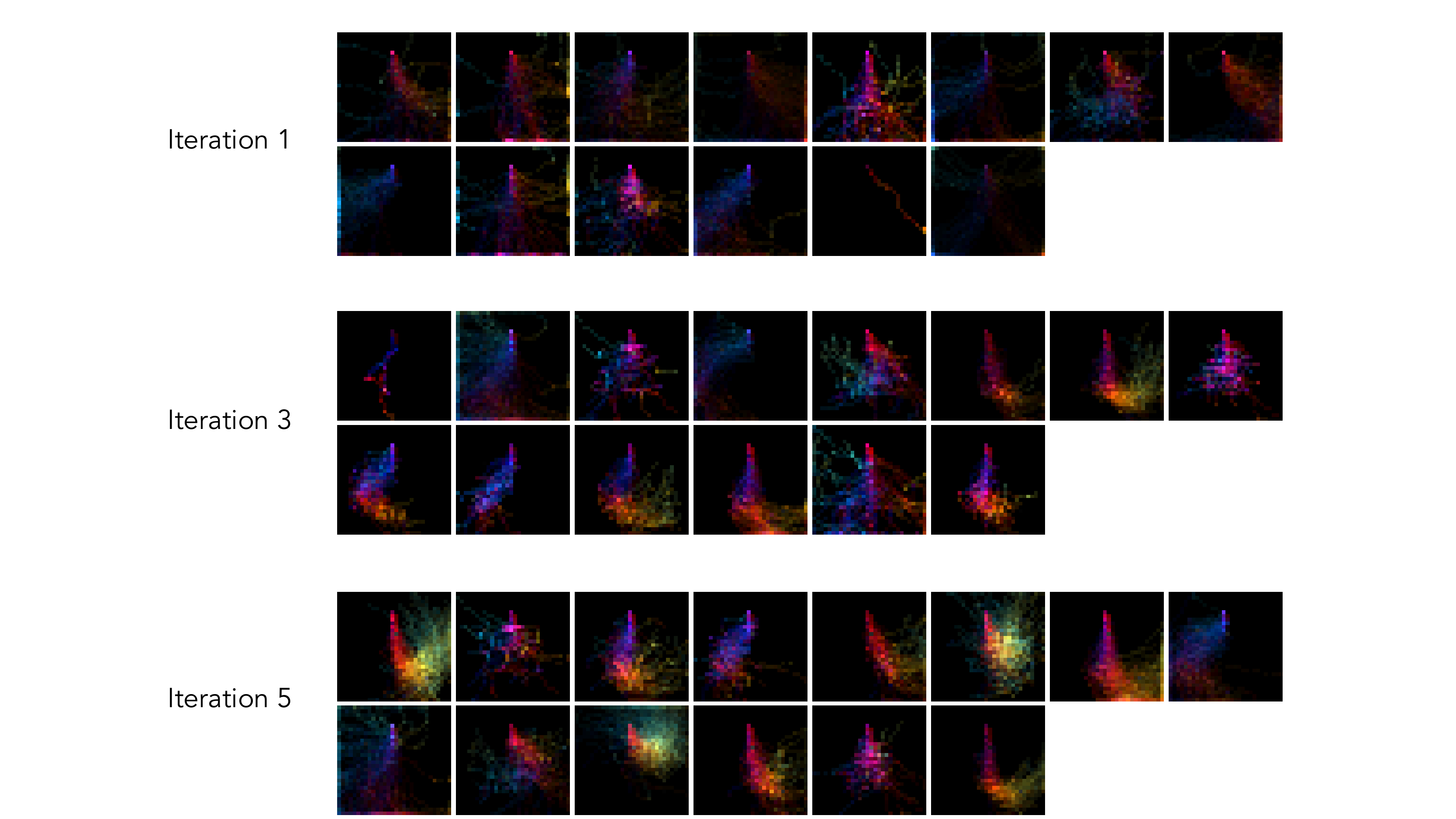}

    \caption{Evolution of the \method{} task distribution over 3 iterations of fitting $q_\phi$ in the random ViZDoom visual navigation environment. We observe evidence of task refinement and incorporation of new tasks.
}
\end{figure}

\end{appendices}

\end{document}